\documentclass[11pt]{article}

\usepackage[final]{acl}

\usepackage{times}
\usepackage{latexsym}
\usepackage{amsfonts}       % Blackboard math symbols
\usepackage{nicefrac}       % Compact symbols for 1/2, etc.
\usepackage{microtype}      % Microtypography
\usepackage{xcolor}         % Colors
\usepackage{bm}             % Bold math symbols
\usepackage[T1]{fontenc}
 \usepackage{amsmath,amssymb,amsthm}
 \usepackage{graphicx} % for \resizebox

\usepackage[utf8]{inputenc}

\usepackage{inconsolata}
\usepackage{booktabs}  %用于绘制专业的三线表 (\toprule, \midrule, \bottomrule)
\usepackage{multirow}  %用于合并行
\usepackage{graphicx}
\usepackage{subcaption}
\usepackage[table,xcdraw]{xcolor} %核心包：用于表格颜色填充
\usepackage{adjustbox}
\usepackage{pifont}

\usepackage{caption}
\usepackage{listings}
\usepackage[most]{tcolorbox}
\tcbuselibrary{listings,skins}

\lstdefinestyle{promptstyle}{
  basicstyle=\ttfamily\scriptsize,
  columns=fullflexible,
  breaklines=true,
  breakatwhitespace=false,
  keepspaces=true,
  showstringspaces=false,
  tabsize=2,
  literate=
    {’}{{'}}1
    {‘}{{`}}1
    {“}{{``}}1
    {”}{{''}}1
    {–}{{--}}1
    {—}{{---}}1
}

\newtcblisting{promptbox}{
  listing engine=listings,
  listing only,
  colback=black!1,
  colframe=black!25,
  boxrule=0.5pt,
  arc=2pt,
  left=6pt,right=6pt,top=6pt,bottom=6pt,
  width=\textwidth,
  listing options={style=promptstyle},
}

\definecolor{rowgray}{gray}{0.92}
\def\wen 1{{\color{red}{\bf [Wen:} {{ 1}}{\bf ]}}}

\title{WavAlign: Enhancing Intelligence and Expressiveness in Spoken Dialogue Models via Adaptive Hybrid Post-Training}

\author{
\textbf{
Yifu Chen\textsuperscript{1}\thanks{These authors contributed equally.}\quad
Shengpeng Ji\textsuperscript{1}\footnotemark[1]\quad
Qian Chen\textsuperscript{2}\footnotemark[1]\quad
Tianle Liang\textsuperscript{1}
}\\
\textbf{
Yangzhuo Li\textsuperscript{1}\quad
Ziqing Wang\textsuperscript{3}\quad
Wen Wang\textsuperscript{2}\quad
Jingyu Lu\textsuperscript{1}\quad
Haoxiao Wang\textsuperscript{1}\quad
Xueyi Pu\textsuperscript{1}
}\\
\textbf{
Fan Zhuo\textsuperscript{1}\quad
Zhou Zhao\textsuperscript{1}\thanks{Corresponding author.}
}\\[0.5em]
$^{1}$~Zhejiang University \quad
$^{2}$~Tongyi Fun Team, Alibaba Group \quad
$^{3}$~Beijing University of Technology \\
\texttt{22551267@zju.edu.cn, zhaozhou@zju.edu.cn}
}

\begin{document}
\maketitle
\begin{abstract}
End-to-end spoken dialogue models have garnered significant attention because they offer a higher potential ceiling in expressiveness and perceptual ability than cascaded systems. However, the intelligence and expressiveness of current open-source spoken dialogue models often remain below expectations. Motivated by the success of online reinforcement learning (RL) in other domains, one might attempt to directly apply preference optimization to spoken dialogue models, yet this transfer is non-trivial. We analyze these obstacles from the perspectives of reward modeling and rollout sampling, focusing on how sparse preference supervision interacts with dense speech generation under shared-parameter updates. Based on the analysis, we propose a modality-aware adaptive post-training recipe that makes RL practical for spoken dialogue: it constrains preference updates to the semantic channel and improves acoustic behavior via explicit anchoring, while dynamically regulating their mixture from rollout statistics to avoid unreliable preference gradients. We evaluate the method across multiple spoken dialogue benchmarks and representative architectures, and observe consistent improvements in semantic quality and speech expressiveness.Our page could be found at \url{https://github.com/MM-Speech/WavAlign}

\end{abstract}

\begin{figure*}[t]
    \centering
    \includegraphics[width=0.8\linewidth]{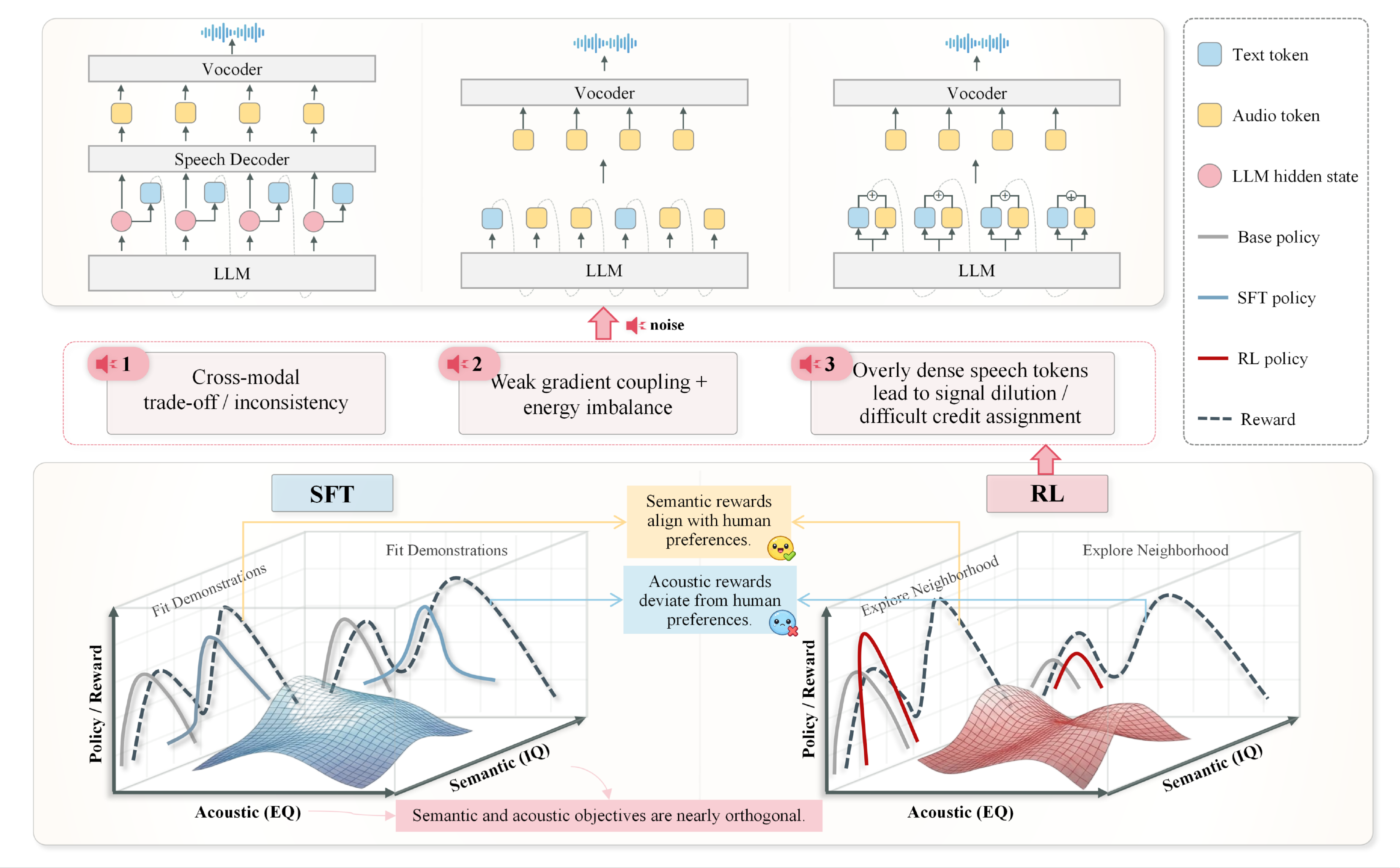}
    \caption{Motivation and failure mode of unified RL for end to end spoken dialogue models.}
    \label{fig:motivation}
    \vspace{-5pt}
\end{figure*}

\section{Introduction}

Spoken dialogue models~\cite{xu2025qwen2,ding2025kimi,wu2025step,fang2024llama,ji2024wavchatsurveyspokendialogue,chen-etal-2025-wavrag} are reshaping human–computer interaction by enabling natural and accessible speech-based interfaces. End-to-end spoken dialogue models directly operate on speech signals and unify speech understanding and generation within a single backbone, allowing joint modeling of semantic content and paralinguistic attributes~\cite{ji2024wavtokenizer,li2026wavbench}. In principle, this paradigm can reduce the error propagation and information loss of cascaded pipelines, while supporting tighter integration between high-level reasoning and fine-grained acoustic expressiveness.

In practice, however, current open-source end-to-end systems still do not consistently surpass strong cascaded baselines, and their semantic capability, naturalness, and expressiveness all leave substantial room for improvement. This gap highlights an open challenge: how to improve semantic dialogue quality and speech naturalness/expressiveness simultaneously within a single end-to-end model, without sacrificing one for the other.

Motivated by the success of Reinforcement Learning from Human Feedback (RLHF) and Reinforcement Learning from AI Feedback (RLAIF) in text and vision~\cite{lee2023rlaif, guo2025deepseek,shen2025vlm}, a natural approach is to apply reinforcement-learning-based preference optimization to end-to-end spoken dialogue. Our empirical findings show that a straightforward, sequence-level preference objective over mixed text–speech outputs is often unreliable for broad, simultaneous gains: semantic preference objectives can improve, yet speech quality frequently degrades, exhibiting acoustic drift and reduced naturalness. Our analysis attributes this instability to an optimization property of omni-modal sequences: preference signals couple weakly across modalities, while the effective gradient energy is highly imbalanced, text gradients dominate shared-parameter updates, and dense speech tokens receive comparatively weak, high-variance supervision. As a result, updates that are beneficial for semantic behavior can inadvertently perturb the delicate acoustic distributions that govern natural prosody and timbre.

Reward modeling further complicates acoustic optimization. Unlike semantic correctness in text, acoustic expressiveness lacks clean, reliable scalar reward signals: rewards are often noisy, underspecified, and entangled with artifacts. When such sparse signals are distributed over long speech token sequences, credit assignment becomes ill-conditioned, and reward hacking can produce speech that scores well while sounding unnatural. These issues are amplified for weaker base models, where high-quality rollouts are rare and preference updates become poorly grounded.

These observations motivate an objective design that separates optimization roles across modalities. Supervised Fine-Tuning (SFT) is effective for constructing and maintaining acoustic feasibility and naturalness, whereas preference optimization is more reliable for semantic refinement, where reward signals are typically more consistent and easier to judge than acoustic expressiveness. Based on this principle, we propose a single-stage adaptive hybrid post-training framework that harmonizes intelligence and expressiveness in one loop: we apply preference optimization only to text tokens to improve semantic behavior, while using SFT as a distribution anchor, which stabilizes the speech token distribution in particular. To mitigate unreliable updates caused by low-quality or low-discriminability rollouts, we further introduce a dynamic gating mechanism that adjusts the balance between supervised and preference-based updates according to rollout validity and training-signal reliability, committing to preference updates only when samples are informative.

Our contributions are summarized as follows:
\begin{itemize}
  \item We identify and characterize key failure modes of unified sequence-level preference optimization for mixed text--speech outputs, including weak cross-modal coupling, gradient-energy imbalance, and noisy acoustic rewards.
  \item We propose a single-stage adaptive hybrid post-training scheme that applies preference optimization to text tokens while anchoring speech tokens with SFT, coupled with a rollout-reliability gating mechanism for stable updates.
  \item Experiments across architectures and benchmarks show consistent gains in both semantic quality and acoustic expressiveness.
\end{itemize}

\section{Related Works}
\noindent\textbf{Reinforcement Learning in Spoken Dialogue Models.}
RL is increasingly used for end-to-end spoken dialogue models, yet prior work typically targets \emph{either} semantic quality (IQ) \emph{or} expressiveness and naturalness (EQ), and joint optimization remains elusive.
Many studies report that optimizing the \emph{full} mixed text--audio token sequence can cause cross-modal instability and text--speech misalignment, prompting decoupled designs such as blocking audio-token gradients~\cite{huang2025step} or adopting text-only objectives~\cite{wu2025aligning}. In parallel, EQ-oriented methods rely on reward modeling and preference data for controllable paralinguistic behavior~\cite{yang2025paras2s,zhang2024speechalign,gao2025emo,lu2026modelingbenchmarkingspokendialogue,ji2025wavrewardspokendialoguemodels}, but can be brittle to reward hacking via spurious acoustic cues~\cite{wang2025rrpo}.
This IQ/EQ separation also motivates modular alternatives that optimize reasoning and speech generation separately~\cite{xu2025qwen3omnitechnicalreport,zheng2025group}. Overall, a unified solution is still missing due to fragmented objectives and persistent cross-modal instability.

\noindent\textbf{Single-Stage Hybrid Post-Training.}
Beyond the conventional two-stage recipe, recent work explores \emph{single-stage} hybrid post-training that mixes SFT and RL within one loop to improve stability, sample efficiency, and capability growth. Methods modulate this trade-off via entropy-aware uncertainty (SRFT~\cite{fu2025srft}), unified feedback continuums (UFT~\cite{liu2025uft}), or dynamic auxiliary terms (CHORD~\cite{zhang2025policy}). Others incorporate reasoning traces with importance sampling (LUFFY~\cite{yan2025learning}) or explicitly optimize synergy and forgetting (ReLIFT~\cite{ma2025learning}, BRIDGE~\cite{chen2025beyond}, MIFO~\cite{yuan2025mitigating}). However, these homogeneous text strategies ill-fit end-to-end spoken dialogue, where scalar preferences are less informative for heterogeneous speech tokens. We therefore propose a modality-aware hybridization principle with a lightweight adaptive controller to regulate hybrid strength based on rollout quality.

\begin{figure*}[t]
\centering
\includegraphics[width=\textwidth]{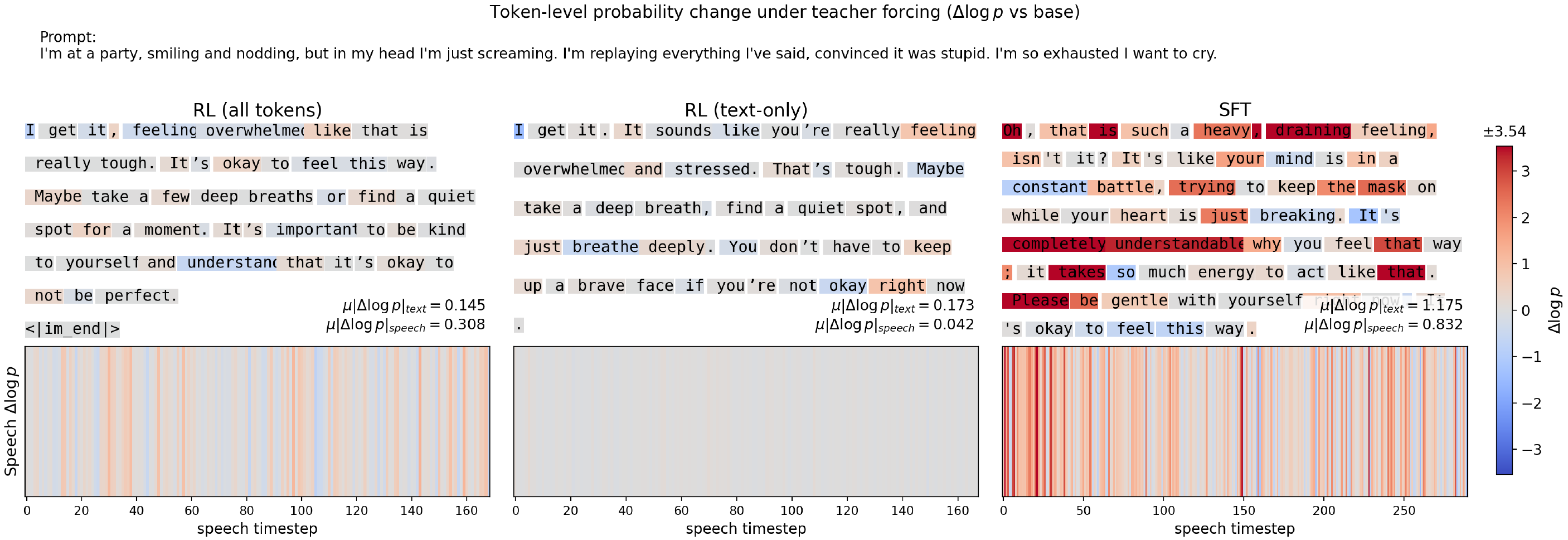} 
\caption{Token-level probability change under teacher forcing ($\Delta \log p$ vs. \ base) for the same prompt.}
\label{fig:logp}
\vspace{-5pt}
\end{figure*}

\section{Methodology}

% =========================================================
% 3.1  Unified Generative Modeling
% =========================================================
\subsection{Preliminaries} 
\subsubsection{Spoken Dialogue Model}
\label{sec:unified_modeling}

We study spoken dialogue models that generate both \emph{text} and \emph{speech} given an input context $x$.
Different architectures realize the generation process differently:
(i) \textbf{Interleaving} generates a single interleaved token stream,
(ii) \textbf{Parallel} generates text and speech streams with coupled states,
and (iii) \textbf{Thinker--Talker} factorizes generation into a ``thinking'' stage and a ``speaking'' stage.
To remain architecture-agnostic, we represent the outputs as two token sequences:
a text sequence $\mathbf{y}^T=(y^T_1,\dots,y^T_L)$ and a speech sequence $\mathbf{y}^S=(y^S_1,\dots,y^S_M)$.
For each emitted token, $c^T_i$ and $c^S_j$ denote its conditioning context under the chosen architecture
(e.g., previous tokens in an interleaved stream, cross-stream hidden states, or a preceding-stage output).

\paragraph{Token-type partition of log-likelihood.}
The model defines a joint conditional distribution $P_\theta(\mathbf{y}^T,\mathbf{y}^S\mid x)$.
Regardless of the internal dependency pattern, the log-likelihood can be partitioned by token types:
\begin{equation}
\label{eq:unified_logprob}
\begin{split}
\log P_\theta(\mathbf{y}^T, &\mathbf{y}^S \mid x) \\
&= \sum_{i=1}^{L} \log P_\theta(y^T_i \mid c^T_i) \\
&\quad + \sum_{j=1}^{M} \log P_\theta(y^S_j \mid c^S_j).
\end{split}
\end{equation}

% =========================================================
% 3.2  Post-training algorithms
% =========================================================
\subsection{Post-Training Algorithms}
\label{sec:post_training}

\subsubsection{Supervised fine-tuning (SFT)}
\label{sec:sft}

Given demonstrations $\mathcal{D}_{\mathrm{sup}}=\{(x,y^{\star})\}$,
SFT minimizes the teacher-forcing cross-entropy:
\begin{equation}
\label{eq:sft}
\begin{split}
\mathcal{L}_{\mathrm{SFT}}&(\theta)
= \\
&-\mathbb{E}_{(x,y^{\star})\sim \mathcal{D}_{\mathrm{sup}}}
\left[
\sum_{t=1}^{|y^{\star}|}
\log \pi_{\theta}\!\left(y^{\star}_t \mid x, y^{\star}_{<t}\right)
\right].
\end{split}
\end{equation}

\noindent\textbf{Property (dense token-level constraint).}
SFT provides a \emph{dense} learning signal at every token position. It is typically the most stable objective.

\subsubsection{Group Relative Policy Optimization}
\label{sec:grpo_full}

For each $x$, GRPO samples a group of $G$ trajectories $\{y^{(i)}\}_{i=1}^{G}$
from a behavior policy $\pi_{\theta_{\mathrm{old}}}(\cdot\mid x)$
and obtains rewards $\{R^{(i)}\}_{i=1}^{G}$ where $R^{(i)}\triangleq R(x,y^{(i)})$.
It uses a group-relative advantage $\hat{A}^{(i)}$ and a PPO-style clipped surrogate with KL regularization:
\begin{equation}
\label{eq:grpo_sketch}
\begin{split}
\mathcal{L}_{\mathrm{GRPO}}&(\theta) = -\,\mathbb{E}\Bigg[ \frac{1}{G}\sum_{i=1}^{G} \sum_{t=1}^{|y^{(i)}|} \min\Bigg( \\
&\quad \rho^{(i)}_{t}\,\hat{A}^{(i)}, \\
&\quad \mathrm{clip}\big(\rho^{(i)}_{t}, 1-\epsilon, 1+\epsilon\big)\,\hat{A}^{(i)}
\Bigg) \Bigg] \\
&+\; \beta\, \mathbb{E}\Big[ \mathrm{KL}\!\big(\pi_{\theta}(\cdot\mid x)\,\|\,\pi_{\mathrm{ref}}(\cdot\mid x)\big) \Big],
\end{split}
\end{equation}
where $\epsilon$ is $\epsilon_{\mathrm{clip}}$ for brevity, and the token-level importance ratio is
$
\rho^{(i)}_{t}
=
\frac{\pi_{\theta}(y^{(i)}_{t}\mid x, y^{(i)}_{<t})}{\pi_{\theta_{\mathrm{old}}}(y^{(i)}_{t}\mid x, y^{(i)}_{<t})}
$.

\noindent\textbf{Property (online; sparse credit shared across tokens).}
GRPO is an \emph{online} method that requires rollouts to obtain rewards.
Although loss are computed at the token level, the advantage $\hat{A}^{(i)}$ is sequence-level (shared across token positions),
which can make credit assignment challenging for long and dense token streams. The KL term acts as a \emph{dense} trust-region constraint that stabilizes optimization and prevents excessive policy drift.

\subsubsection{Offline DPO-family}
\label{sec:dpo_full}

Given pairwise preference data $\mathcal{D}_{\mathrm{pref}}=\{(x,y^{+},y^{-})\}$,
DPO optimizes a logistic loss on the reference-corrected log-ratio gap.
We define
\begin{equation}
\label{eq:dpo_delta}
\begin{split}
\Delta(x,&y^{+},y^{-};\theta) \\
&= \Big(\log \pi_{\theta}(y^{+}\mid x) - \log \pi_{\theta}(y^{-}\mid x)\Big) \\
&\quad - \Big(\log \pi_{\mathrm{ref}}(y^{+}\mid x) - \log \pi_{\mathrm{ref}}(y^{-}\mid x)\Big),
\end{split}
\end{equation}
and minimize
\begin{equation}
\label{eq:dpo_loss}
\begin{split}
\mathcal{L}_{\mathrm{DPO}}&(\theta) = \\
&-\mathbb{E}_{\mathcal{D}_{\mathrm{pref}}}
\Big[
\log \sigma\!\big(\gamma\cdot \Delta(x,y^{+},y^{-};\theta)\big)
\Big],
\end{split}
\end{equation}
where $\sigma(\cdot)$ is the logistic sigmoid and $\gamma>0$ is a temperature.

\noindent\textbf{Property (offline; sparse preference supervision).}
DPO is \emph{offline} and does not require rollouts during optimization, making it scalable in practice.
Supervision signal comes only from pairwise preferences and performance depends on preference quality/coverage and potential judge bias.

\paragraph{Token-subset restricted likelihood.}
In mixed-modality settings, it is sometimes useful to restrict preference-driven updates to a subset of token positions.
Given an index set $\mathcal{M}(y)\subseteq \{1,\ldots,|y|\}$, define the masked score:
\begin{equation}
\label{eq:masked_score}
s_{\mathcal{M}}(x,y;\theta)
\triangleq
\sum_{t\in \mathcal{M}(y)}
\log \pi_{\theta}\!\left(y_t \mid x, y_{<t}\right).
\end{equation}
Replacing $\log \pi_\theta(y\mid x)$ with $s_{\mathcal{M}}(x,y;\theta)$ yields token-subset restricted variants of
SFT/PO objectives, enabling explicit control over which token types receive preference gradients.

\begin{figure*}[t]
    \centering
    \includegraphics[width=0.7\linewidth]{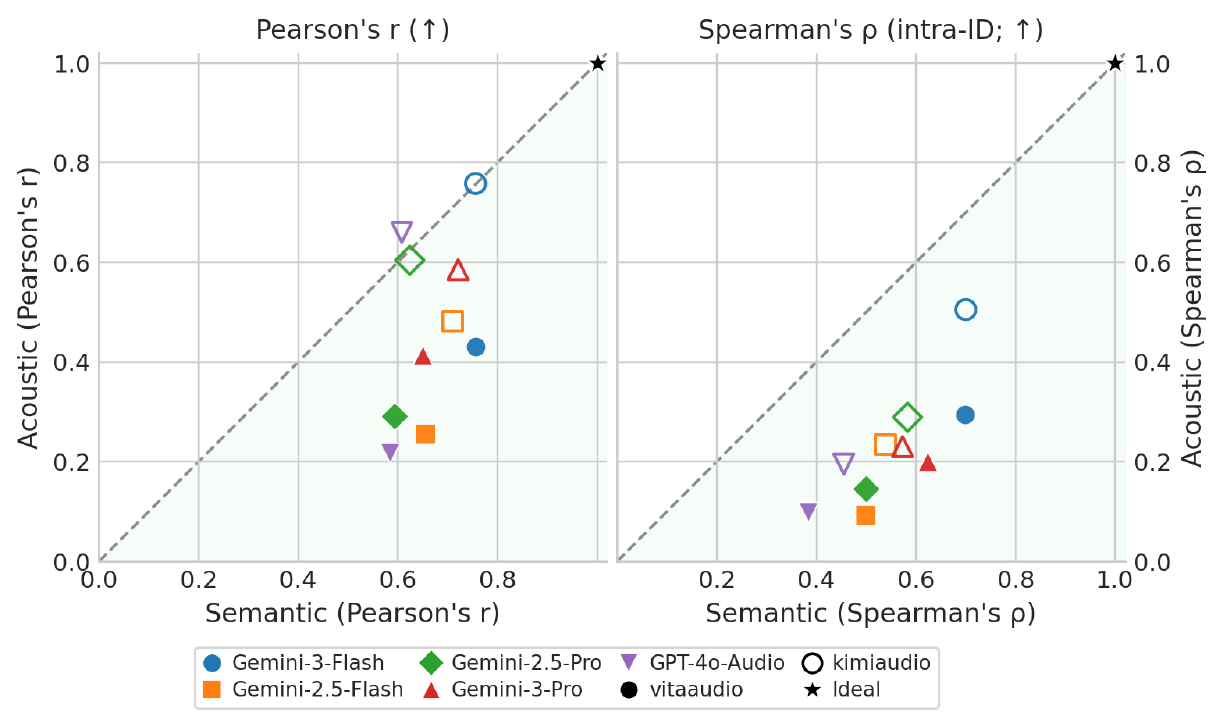}

    \caption{Judge/reward-model agreement with human evaluation on semantic vs. acoustic dimensions.}
    \label{fig:consistency}
\end{figure*}

\begin{figure*}[t]
    \centering
    \includegraphics[width=0.7\linewidth]{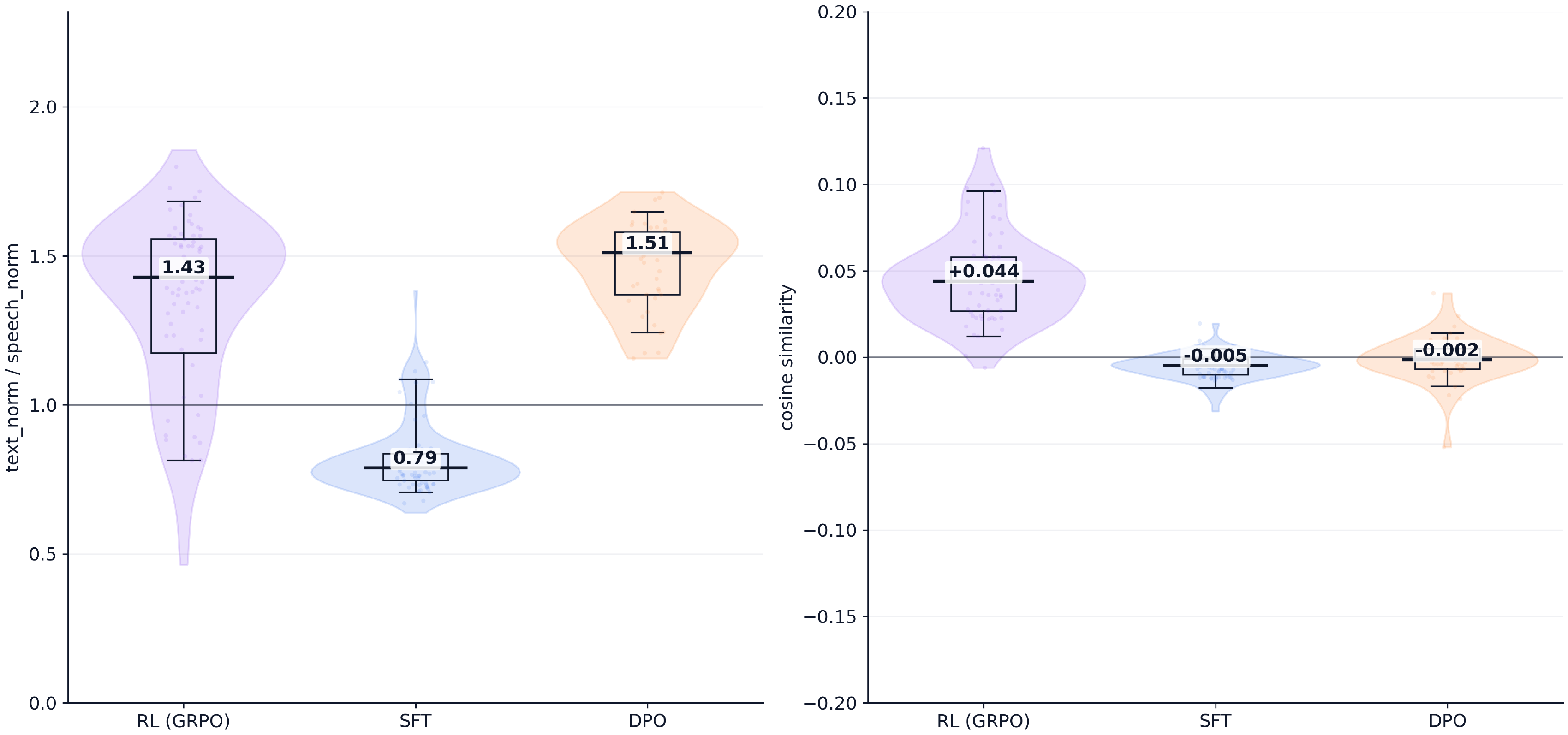}
    \caption{Empirical geometry of text vs. speech gradients under different objectives.}
    \label{fig:grad}
    \vspace{-5pt}
\end{figure*}

\begin{figure*}[t]
    \centering
    \includegraphics[width=0.7\linewidth]{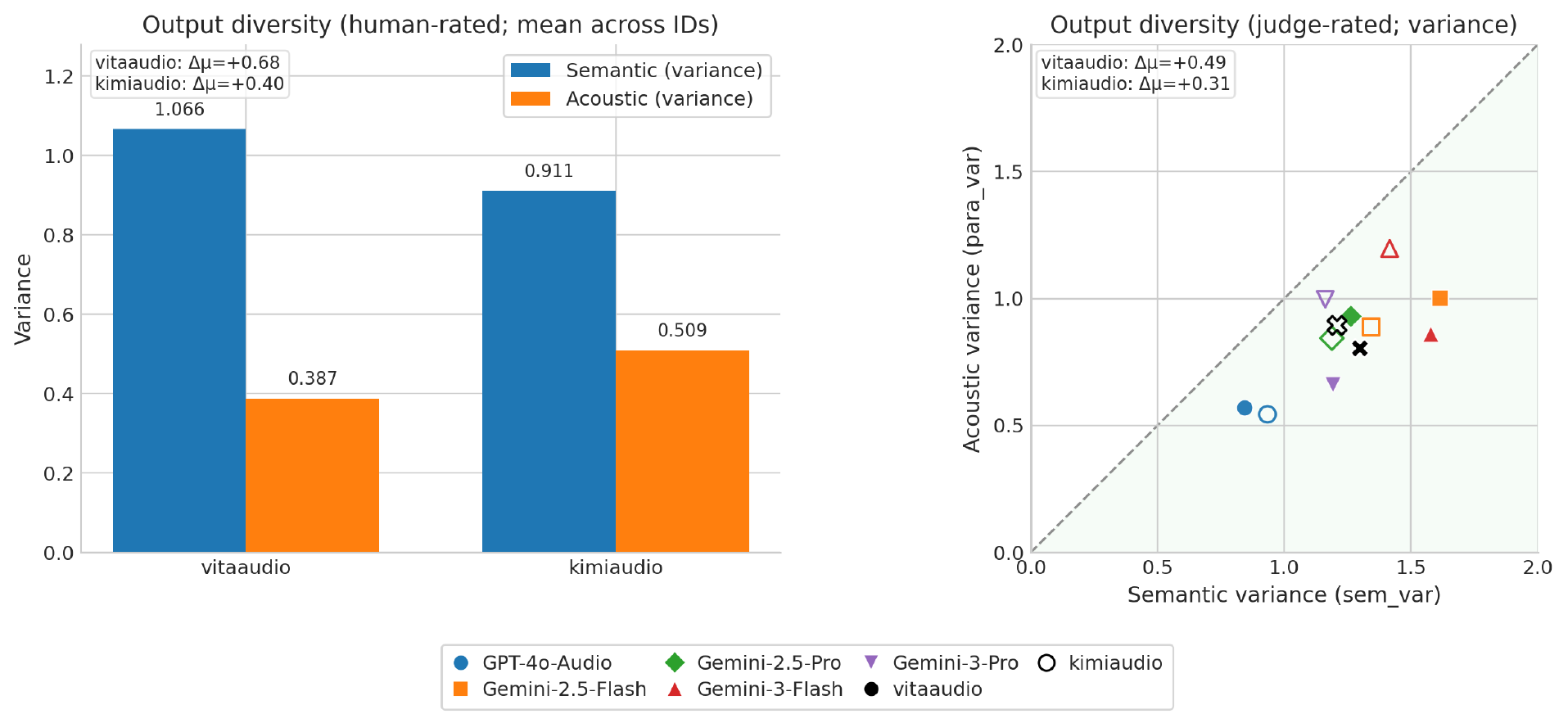}
    \caption{Semantic vs. acoustic diversity under repeated sampling reveals weaker acoustic discriminability.}
    \label{fig:diversity}
    \vspace{-5pt}
\end{figure*}

\subsection{Observations}
\label{sec:motivation}
As highlighted in Fig.~\ref{fig:motivation}, applying a single unified RL/PO objective to the mixed text--speech output leads to three coupled issues: (i) cross-modal trade-off and inconsistency, (ii) weak cross-modal gradient coupling with severe energy imbalance, and (iii) reward/signal dilution when sparse feedback is spread over overly dense speech tokens, making credit assignment ill-posed.
Under token-score--based objectives, the additivity in Eq.~(1) yields a natural gradient split
\begin{equation}
\nabla_\theta L(\theta) \;=\; \nabla_\theta L^{(T)}(\theta) \;+\; \nabla_\theta L^{(S)}(\theta),
\end{equation}
which exposes the mechanism behind Fig.~\ref{fig:motivation}: semantic (text) updates carry much higher effective energy, while the acoustic component receives weakly informative, high-variance signals due to near-orthogonal coupling, whose \emph{accumulated effect over dense speech tokens} can destabilize prosody/timbre.
This motivates a division of labor: restrict preference-driven updates to $I_T$ for semantic refinement, and use dense supervision to anchor $I_S$ to preserve speech naturalness and expressive stability.

However, even after restricting preference-driven updates to $I_T$, where preference signals are typically most informative, preference optimization alone is insufficient in our setting due to two structural bottlenecks.

\begin{figure*}[t]
    \centering
    \includegraphics[width=0.8\linewidth]{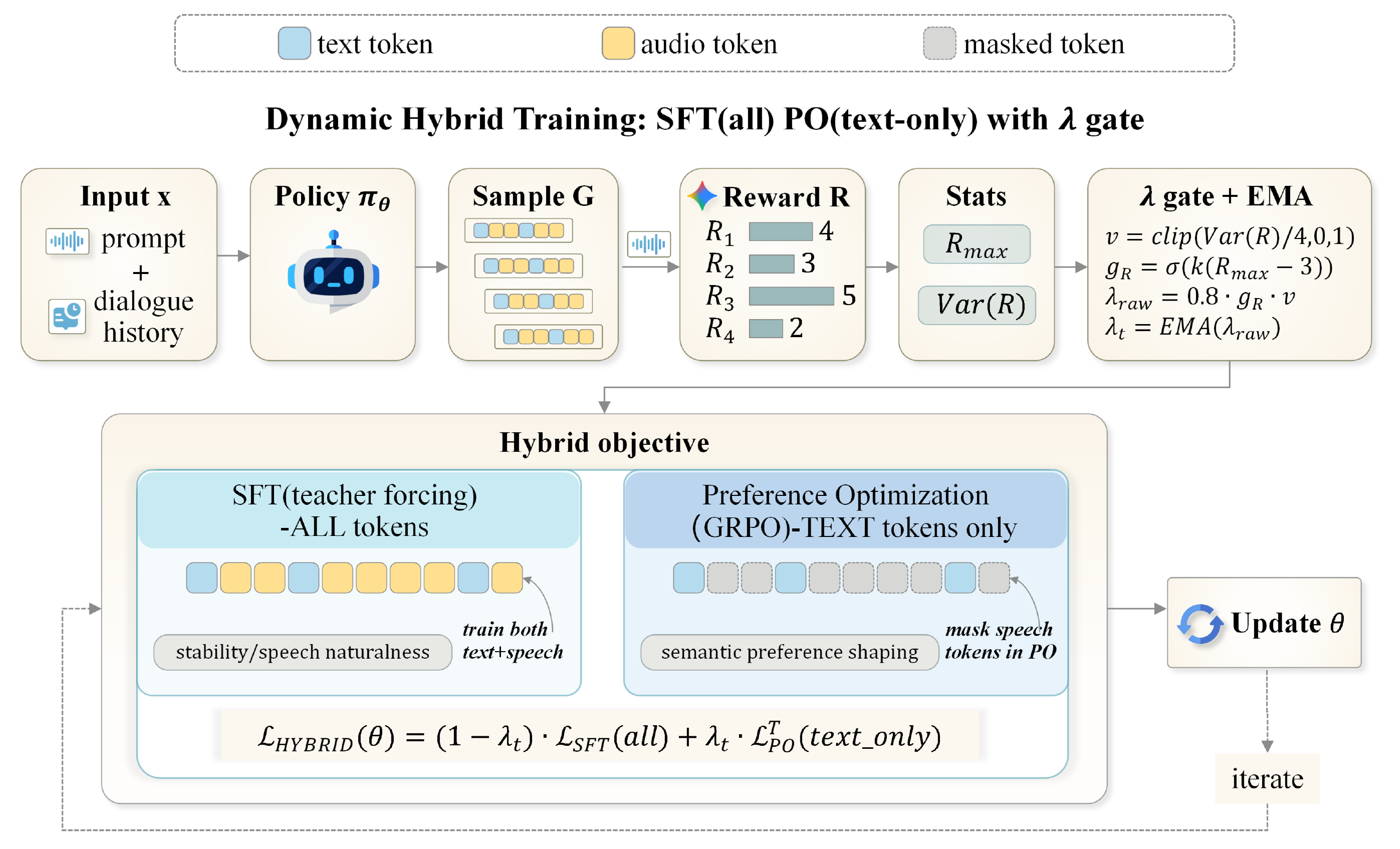}
    \caption{Overview of the proposed single stage adaptive hybrid post training.}
    \label{fig:method}
    \vspace{-5pt}
\end{figure*}

\noindent \textbf{(Bottleneck I)} Practical PO is inherently \emph{local} under stability constraints (e.g., clipping and/or reference penalties), typically inducing substantially smaller distributional shifts than supervised fine-tuning (SFT), and may fail to move the model out of suboptimal regions when rollouts provide limited improvement signals.

\noindent \textbf{(Bottleneck II)} Speech naturalness/expressiveness lacks reliably learnable preference signals and strong rollouts: acoustic reward judgments can be noisy or misaligned with human preference, and weak base models rarely sample high-quality acoustic trajectories, yielding low reward discrimination, which makes direct preference learning on dense acoustic decisions fragile or even harmful.
In what follows, we present a set of empirical observations (Figs.~\ref{fig:logp}--\ref{fig:diversity}) and distill them into design conclusions that motivate the dynamic hybrid objective in Sec.~\ref{sec:hybrid_obj}.

\paragraph{Observation 1 (Fig.~\ref{fig:logp}): SFT yields larger, coherent distribution shifts, while stabilized PO/RL is typically local.}
Fig.~\ref{fig:logp} shows that SFT induces substantially larger and more consistent token-level probability changes across the sequence, whereas PO/RL updates are smaller and localized under trust-region--like stability constraints. This matches Bottleneck~I: on-policy rollouts often provide limited improvement signal, making PO a local shaper that may not escape suboptimal regions. Details of the teacher-forcing probability-change metric and modality partition are in Appendix~\ref{sec:appendix_distribution}. \emph{Implication:} use SFT to enact reliable global shifts/anchoring, and PO/RL to refine behavior locally when preference signals are informative.

\paragraph{Observation 2 (Fig.~\ref{fig:consistency}): Preference/reward is more informative for semantics than acoustics.}
Across repeated samples (Fig.~\ref{fig:consistency}), reward-model judgments agree with humans more strongly and stably on semantic quality than on acoustic quality, where agreement is weaker and more variable. Full human-eval protocol, agreement tables, and judge/reward prompts are in Appendix~\ref{sec:appendix_rm_consistency} and Appendix~\ref{sec:appendix_reward_prompts}. \emph{Implication:} applying high-variance PO over dense speech tokens is fragile (noise amplification and harder credit assignment), so preference-driven updates should focus on $I_T$, while dense supervision stabilizes $I_S$.

\paragraph{Observation 3 (Fig.~\ref{fig:grad}): Preference gradients concentrate on semantics; full-token PO yields low-SNR, high-variance updates on dense acoustics.}
Fig.~\ref{fig:grad} indicates weak cross-modal coupling (near-zero expected cosine similarity with high variance) and that preference objectives allocate most gradient energy to the semantic component. As a result, applying PO to the full mixed token stream assigns the same sequence-level credit to a large number of speech token decisions that are only weakly correlated with the preference signal, producing near-zero-mean but high-variance acoustic gradients. The accumulation of these noisy acoustic updates can destabilize prosody/timbre, motivating preference updates on token subsets. Gradient decomposition and all-layer analysis details are in Appendix~\ref{sec:appendix_grad}.

\paragraph{Observation 4 (Fig.~\ref{fig:diversity}): Rollout discriminability is uneven and stage-dependent, favoring adaptive gating over fixed mixing.}
With weaker base models, rollouts rarely contain high-quality acoustic trajectories, yielding low reward discrimination (small variance, few high-score samples); Fig.~\ref{fig:diversity} further shows diversity/variance is uneven (often weakest along acoustics) and changes over training stages. Statistics and definitions are in Appendix~\ref{sec:appendix_diversity}. \emph{Implication:} a fixed hybrid weight either applies PO when signals are weak/noisy (instability) or underuses PO when discriminative samples exist; thus an adaptive controller (Sec.~\ref{sec:hybrid_obj}) should modulate $\lambda_t$ using normalized reward variance and a good-sample existence gate.

\paragraph{Summary.}
Taken together, these observations motivate a principled division of labor:
\textbf{SFT} acts as a robust distribution-shifting and feasibility-building operator (especially for speech naturalness/expressiveness through dense supervision),
while \textbf{PO} acts as a local preference-shaping operator that refines semantic dialogue behavior when the preference signal is informative.
Sec.~\ref{sec:hybrid_obj} formalizes this view into a lightweight dynamic hybrid objective with adaptive gating.

\subsection{Dynamic Hybrid Post-Training Objective}
\label{sec:hybrid_obj}

Fig.~\ref{fig:method} summarizes our single-stage dynamic hybrid post-training loop.
At each step, we sample a group of $G$ spoken replies from $\pi_\theta$; importantly, we decode the generated speech into audio and \emph{feed the model's spoken reply directly to a reward model} to obtain scalar rewards $R_t=\{r_{t,i}\}_{i=1}^{G}$.
Motivated by Fig.~\ref{fig:motivation}, we keep SFT as an explicit distribution anchor for acoustic stability, while applying preference optimization only on text tokens to refine semantics.
We thus optimize
\begin{equation}
\mathcal{L}_{\text{hybrid}}(\theta)
=
(1-\lambda_t)\,\mathcal{L}_{\text{SFT}}(\theta)
+
\lambda_t\,\mathcal{L}^{(T)}_{\text{GRPO}}(\theta),
\label{eq:hybrid}
\end{equation}
where $\mathcal{L}^{(T)}_{\text{GRPO}}$ masks the score in Eq.~\ref{eq:masked_score} with $M(y)=I_T(y)$, i.e., preference gradients are restricted to text tokens.

% === Table 1: Main Results IQ (Categorized) ===
\begin{table*}[t]
\centering
\small
\resizebox{\textwidth}{!}{
    \begin{tabular}{lccccccccc|c|ccccc|c}
    \toprule
    \multirow{2}{*}{\textbf{Method}} & \multicolumn{9}{c|}{\textbf{VoiceBench}} & \multirow{2}{*}{\textbf{Avg}} & \multicolumn{5}{c|}{\textbf{OpenAudioBench}} & \multirow{2}{*}{\textbf{Avg}} \\
    \cmidrule(lr){2-10} \cmidrule(lr){12-16}
     & \textbf{Alpaca} & \textbf{Common} & \textbf{Wild} & \textbf{SD-QA} & \textbf{MMSU} & \textbf{OBQA} & \textbf{BBH} & \textbf{IFEval} & \textbf{Adv} & & \textbf{Alpaca} & \textbf{Llama} & \textbf{Reason} & \textbf{Trivial} & \textbf{Web} & \\
    \midrule
    \multicolumn{17}{c}{\cellcolor{gray!20}\textbf{VITA Architecture (Interleaved)}} \\
    \midrule
    \textcolor{gray}{VITA-Base} & 3.83 & 3.44 & 3.09 & 29.2 & 48.7 & 74.3 & 58.2 & 26.2 & 94.1 & -- & 60.6 & 73.8 & 44.2 & 42.9 & 53.5 & 55.0 \\
    SFT (Teacher Forcing) & 3.45 & 3.12 & 2.85 & 27.6 & 45.1 & 71.5 & 54.9 & 28.4 & \textbf{99.2} & -- & 55.6 & 71.1 & 38.4 & 39.9 & 48.3 & 50.7 \\
    \midrule
    \multicolumn{17}{l}{\textit{DPO Baselines}} \\
    Full-Token DPO & 3.60 & 3.29 & 2.89 & 30.2 & 44.7 & 69.2 & 56.8 & 22.6 & 65.0 & -- & 20.1 & 55.4 & 33.4 & 29.8 & 36.6 & 35.1 \\
    Text-Token DPO & 3.91 & 3.32 & 3.13 & 31.1 & 45.6 & 69.7 & \textbf{60.3} & \textbf{32.8} & 71.3 & -- & 57.2 & \underline{74.3} & 43.1 & 43.1 & \underline{54.3} & 54.4 \\
    \midrule
    \multicolumn{17}{l}{\textit{RL Baselines}} \\
    Full-Token RL (Unified) & 4.03 & \underline{3.45} & 3.19 & 29.9 & 49.0 & 74.1 & 55.6 & 29.4 & 96.3 & -- & 63.8 & 73.3 & 43.7 & \underline{43.3} & 52.8 & 55.4 \\
    Text-Token RL (Unified) & \underline{4.09} & 3.44 & \underline{3.20} & \underline{31.3} & \underline{50.0} & \underline{75.4} & 56.7 & 30.2 & 96.3 & -- & \underline{64.6} & \textbf{74.6} & \underline{44.4} & \textbf{44.4} & 53.2 & \underline{56.2} \\
    SFT + RL (Two-Stage) & 3.49 & 3.32 & 2.69 & 22.5 & 44.7 & 70.8 & 54.2 & 25.5 & \underline{98.8} & -- & 54.0 & 66.2 & 32.8 & 35.1 & 49.0 & 47.4 \\
    \midrule
    \rowcolor{rowgray} \textbf{Ours (Dynamic)} & \textbf{4.22} & \textbf{3.51} & \textbf{3.29} & \textbf{31.5} & \textbf{51.4} & \textbf{77.1} & \underline{59.9} & \underline{32.5} & 97.1 & -- & \textbf{68.4} & \textbf{74.6} & \textbf{46.1} & \textbf{44.4} & \textbf{54.7} & \textbf{57.6} \\

    \midrule
    \multicolumn{17}{c}{\cellcolor{gray!20}\textbf{KimiAudio Architecture (Parallel)}} \\
    \midrule
    KimiAudio-Base & 4.46 & 3.97 & 3.42 & 63.1 & 62.2 & 83.5 & 64.2 & 61.1 & \textbf{100.0} & -- & 75.7 & \underline{79.3} & 58.0 & \textbf{62.1} & 70.2 & 69.1 \\
    SFT & 4.15 & 3.65 & 3.10 & 59.8 & 58.4 & 79.5 & 61.2 & \underline{64.5} & \textbf{100.0} & -- & 71.4 & 75.2 & 52.8 & 58.4 & 66.9 & 64.9 \\
    \midrule
    \multicolumn{17}{l}{\textit{Preference Optimization (DPO \& RL)}} \\
    Full-Token DPO & 4.05 & 3.60 & 3.05 & 58.2 & 55.1 & 76.8 & 59.4 & 58.4 & 88.5 & -- & 68.2 & 70.4 & 50.1 & 55.3 & 65.1 & 61.8 \\
    Full-Token RL & \underline{4.52} & \underline{4.05} & \underline{3.50} & \underline{65.2} & \underline{63.8} & \underline{84.6} & \underline{64.8} & 62.8 & \textbf{100.0} & -- & \underline{75.8} & 78.5 & \underline{58.8} & 61.2 & \textbf{71.5} & \underline{69.2} \\
    \midrule
    \rowcolor{rowgray} \textbf{Ours (Dynamic)} & \textbf{4.58} & \textbf{4.22} & \textbf{3.68} & \textbf{67.9} & \textbf{66.5} & \textbf{87.1} & \textbf{68.3} & \textbf{66.8} & \underline{99.5} & -- & \textbf{78.5} & \textbf{81.2} & \textbf{61.5} & \underline{61.8} & \underline{71.1} & \textbf{70.8} \\
    \bottomrule
    \end{tabular}
}
\caption{\textbf{Main Results on Intelligence (IQ).} Best results are \textbf{bolded} and second best are \underline{underlined}. Avg for VoiceBench is omitted (mixed scales); Avg for OpenAudioBench is the arithmetic mean of the five sub-task scores.}
\vspace{-5pt}
\label{tab:main_iq}
\end{table*}

\paragraph{Lightweight gating for $\lambda_t$.}
We increase $\lambda_t$ only when rollouts are (i) \emph{directionally reliable} (at least one acceptable sample exists) and (ii) \emph{discriminative} (candidates are well-separated), matching the two statistics shown in Fig.~\ref{fig:method}.
Let $R_{\max,t}=\max(R_t)$ and define a normalized variance
\begin{equation}
v_t \triangleq \mathrm{clip}\!\left(\frac{\mathrm{Var}(R_t)}{4},\,0,\,1\right),
\end{equation}
where $4$ is the maximum variance on a 1--5 Likert scale (bimodal at $\{1,5\}$), so $v_t\!\in[0,1]$ is comparable across steps.
The direction gate is
\begin{equation}
g_t(R)=\sigma\!\left(k\,(R_{\max,t}-3)\right),
\end{equation}
where the threshold $3$ corresponds to neutral/acceptable quality; if all samples fall below $3$, preference gradients are typically noisy/misdirected. We use $g_t(v_t)=v_t$ as the information gate (where $v_t$ is the normalized reward variance defined above) and set
\begin{equation}
\lambda_t^{\text{raw}}=\lambda_{\max}\,g_t(R)\,g_t(V), \qquad \lambda_{\max}=0.8,
\end{equation}
so that at least $1-\lambda_{\max}=0.2$ of SFT is always retained as a safety anchor against acoustic drift when rewards are imperfect. The slope $k$ is the only sharpness hyperparameter, controlling how softly we transition from ``mostly SFT'' to ``more GRPO''.

\paragraph{EMA smoothing.}
To reduce step-to-step oscillations from on-policy sampling, we smooth the weight itself:
\begin{equation}
\lambda_t=(1-\alpha)\lambda_t^{\text{raw}}+\alpha \lambda_{t-1},
\qquad \alpha=0.9,
\end{equation}
where $\alpha=0.9$ provides a strong low-pass filter that stabilizes training while remaining responsive to sustained improvements in rollout quality/discriminability.

\section{Experiments}
\label{sec:experiments}

\subsection{Experimental Setup}

\paragraph{Training Data.}
To cover both \emph{intelligence} (reasoning, knowledge, instruction following, safety) and \emph{expressiveness} (paralinguistic controllability and empathy), we curate a mixed training set of 13.5k audio-instruction samples from public and self-constructed sources, including UltraChat~\cite{ding2023enhancing}, SciQ~\cite{SciQ}, GSM8K~\cite{cobbe2021gsm8k}, SHP~\cite{pmlr-v162-ethayarajh22a}, ExamQA, Alpaca~\cite{alpaca}, ScienceQA~\cite{lu2022learn}, Ai2ARC~\cite{allenai:arc}, PKUSafe~\cite{ji2024pku}, as well as self-constructed logic and expressiveness data. All 13.5K training samples are from public or self-constructed sources; no proprietary in-house data are used. We further build preference pairs via repeated sampling and judge-based scoring to support offline preference learning. The detailed data composition, construction procedures, and preference-pair pipeline are provided in Appendix~\ref{sec:appendix_train_datasets}.

\paragraph{Benchmarks and Metrics.}
We evaluate on three benchmarks comprising \textbf{18 sub-tasks} to jointly assess semantic competence (IQ) and expressive capability (EQ).
\textbf{VoiceBench} evaluates instruction following and safety, including \textit{AlpacaEval, CommonEval, WildVoice, SD-QA, MMSU, OBQA, BBH, IFEval,} and \textit{AdvBench}.
\textbf{OpenAudioBench} focuses on knowledge and reasoning with \textit{Alpaca, Llama, Web, Trivial,} and \textit{Reason}.
\textbf{VStyle} targets paralinguistic control via \textit{Acoustic Attributes, Instruction Following (Style), Role Play,} and \textit{Empathy}.
We strictly follow the official evaluation protocol for each benchmark: VoiceBench and OpenAudioBench are evaluated on text outputs via their official pipelines (GPT-4o-mini and GPT-4o as judges, respectively); VStyle is evaluated on speech outputs via its official procedure using Gemini-2.5-Pro. All scores are computed using the official scripts and judges for each benchmark.

\paragraph{Baselines}
To demonstrate architectural generality, we experiment with two distinct end-to-end speech dialogue backbones:
(i) \textbf{VITA-Audio}~\cite{long2025vitaaudiofastinterleavedcrossmodal}, which emits an interleaved sequence of text and speech tokens, and
(ii) \textbf{KimiAudio}~\cite{ding2025kimi}, which follows a parallel design.
We group baselines into:
\textbf{(1) Standard:} the \textit{Base Model} and \textit{SFT} (teacher forcing).
\textbf{(2) DPO:} offline Direct Preference Optimization applied to either \textit{Full} tokens or \textit{Text} tokens only.
\textbf{(3) RL:} GRPO applied to \textit{Full Tokens} or \textit{Text Tokens} only, plus a sequential \textit{Two-Stage} recipe (SFT$\rightarrow$RL).
For DPO baselines, preference supervision is constructed via repeated sampling and judge scoring; details are in Appendix~\ref{sec:appendix_train_datasets}.

\paragraph{Implementation details.}
All training runs use 4$\times$A100 GPUs.
For RL, we adopt a KL-regularized objective with coefficients $\beta_{\text{text}}=0.01$ and $\beta_{\text{speech}}=0.01$.
Unless otherwise specified, we use group size $G=4$, sampling temperature $T=0.9$, top-$p=0.9$, learning rate $1\mathrm{e}{-6}$, batch size $1$, and maximum sequence length $2048$.
We use Gemini-2.5-pro as the reward model to score model speech responses; prompting templates and output formats for semantic and paralinguistic scoring are given in Appendix~\ref{sec:appendix_reward_prompts}.

\subsection{Main Results}
\paragraph{Intelligence (IQ).}
Table~1 shows that teacher-forced SFT \emph{consistently underperforms} the base model on reasoning-heavy subsets, suggesting an alignment tax in our audio-instruction setting.
We attribute this mainly to the \emph{broad multi-domain} supervision: a comparatively small audio-instruction corpus must cover heterogeneous skills, which increases gradient interference and can overwrite pre-trained reasoning behaviors.
Across both backbones, full-token preference optimization is suboptimal, while restricting preference updates to \emph{text/semantic tokens} yields more reliable IQ gains, supporting modality-decoupled optimization.
Building on this, our dynamic hybrid further mitigates catastrophic forgetting while preserving preference-learning benefits, achieving the strongest overall IQ among compared methods.

\paragraph{Expressiveness (EQ).}
On VStyle (Table~2), SFT remains highly competitive, especially on style instruction following and acoustic attributes, indicating that dense supervision is effective for imprinting fine-grained paralinguistic realizations.
In contrast, naive preference optimization over the full mixed-modality sequence can be unstable for expressive speech: full-token DPO exhibits severe degradation, consistent with noisy or weakly discriminative acoustic reward signals.
Our method achieves the best aggregate EQ across dimensions on both architectures, while staying close to the best style-following score, yielding a better IQ--EQ Pareto trade-off than either component alone.

% === Table 2: Main Results EQ (Categorized) ===
\begin{table}[t]
\centering
\small
\resizebox{\columnwidth}{!}{
    \begin{tabular}{lccccc}
    \toprule
    \textbf{Method} & \textbf{Acoustic} & \textbf{Instruct.} & \textbf{Role Play} & \textbf{Empathy} & \textbf{Avg} \\
    \midrule
    \multicolumn{6}{c}{\cellcolor{gray!20}\textbf{VITA Architecture}} \\
    \midrule
    VITA-Base & 2.26 & 1.76 & 2.15 & 4.01 & 2.55 \\
    SFT (Teacher Forcing) & \underline{2.34} & \textbf{2.29} & \underline{2.31} & 3.42 & 2.59 \\
    \midrule
    \textit{DPO Baselines} & & & & & \\
    Full-Token DPO & 1.49 & 1.25 & 1.10 & 1.05 & 1.22 \\
    Text-Token DPO & 2.03 & 1.64 & 2.19 & \underline{4.38} & 2.56 \\
    \midrule
    \textit{RL Baselines} & & & & & \\
    Full-Token RL (Unified) & 2.16 & 1.64 & 1.97 & 3.95 & 2.43 \\
    Text-Token RL (Unified) & 2.21 & \underline{1.93} & 2.08 & 4.02 & 2.56 \\
    \midrule
    \rowcolor{rowgray} \textbf{Ours (Dynamic)} & \textbf{2.55} & \underline{2.25} & \textbf{2.41} & \textbf{4.44} & \textbf{2.91} \\

    \midrule
    \multicolumn{6}{c}{\cellcolor{gray!20}\textbf{KimiAudio Architecture}} \\
    \midrule
    KimiAudio-Base & 2.53 & 2.31 & 1.73 & 3.67 & 2.56 \\
    SFT & \underline{2.65} & \textbf{2.58} & \underline{1.95} & 3.65 & 2.71 \\
    \midrule
    \textit{Preference Opt.} & & & & & \\
    Full-Token DPO & 1.85 & 1.55 & 1.30 & 2.10 & 1.70 \\
    Full-Token RL & 2.58 & 2.25 & 1.88 & \underline{3.88} & 2.65 \\
    \midrule
    \rowcolor{rowgray} \textbf{Ours (Dynamic)} & \textbf{2.78} & \underline{2.52} & \textbf{2.15} & \textbf{4.15} & \textbf{2.90} \\
    \bottomrule
    \end{tabular}
}
\caption{\textbf{Main Results on Expressiveness (EQ).} Comparison on VStyle. Avg is the arithmetic mean of the four dimension scores. \textbf{Bold}/\underline{Underline} indicate best/second best.}
\vspace{-5pt}
\label{tab:main_eq}
\end{table}

\subsection{Ablation Studies and Analysis}
\label{sec:ablation}

\subsubsection{Weighting schemes and optimization scope}
\label{sec:ablation_weighting}
We study how to combine dense supervision (SFT) with preference optimization (RL/PO) for mixed-modality spoken outputs by varying two factors: \textbf{(i) optimization scope: }applying preference updates to \emph{all} tokens versus restricting them to \emph{text} tokens and \textbf{(ii) weighting strategy: }\emph{fixed} SFT/RL mixtures versus \emph{dynamic} weights predicted from rollout quality. All variants share the same backbone, data, and training budget; we report \textbf{IQ} (mean over VoiceBench reasoning subsets: MMSU/OBQA/BBH/IFEval) and \textbf{EQ} (average VStyle score).
Table~\ref{tab:ablation_weighting} indicates that \emph{scope} is crucial: with the same fixed 0.5/0.5 mixture, applying preference optimization only to text tokens clearly outperforms updating all tokens (52.60/2.60 vs.\ 48.70/2.48 in IQ/EQ), implying preference gradients are most effective when focused on semantic-bearing regions. Fixed weights also reveal an IQ--EQ trade-off—favoring SFT (0.7/0.3) improves EQ (2.72) but lowers IQ (49.94). Dynamic weighting over all tokens remains limited (48.84/2.50), whereas dynamic gating with text-token scope delivers the best overall result; EMA smoothing is important for stability (w/o EMA: 53.15/2.53 vs.\ ours: 55.24/2.92).

\begin{table}[t]
\centering
\small
\setlength{\tabcolsep}{6pt}
\resizebox{\columnwidth}{!}{
\begin{tabular}{llcc}
\toprule
\textbf{Scope} & \textbf{Strategy} & \textbf{IQ} & \textbf{EQ} \\
\midrule
All Tokens & Fixed Weights (0.5/0.5) & 48.70 & 2.48\\
Text Tokens & Fixed Weights (0.5/0.5) & 52.60 & 2.60 \\
Text Tokens & Fixed Weights (0.7 SFT / 0.3 RL) & 49.94 & 2.72 \\
All Tokens & Dynamic Weights & 48.84 & 2.50 \\
Text Tokens & Dynamic Weights w/o EMA & 53.15 & 2.53 \\
\rowcolor{gray!12}
Text Tokens & \textbf{Ours (Dynamic Weights)} & \textbf{55.24} & \textbf{2.92} \\
\bottomrule
\end{tabular}
}
\caption{Weighting schemes and optimization scope. \emph{Scope} refers to the token subset over which the GRPO/RL loss is applied; SFT always covers all tokens regardless of scope.}
\label{tab:ablation_weighting}
\end{table}

\begin{table}[t]
    \centering
    \small
    \setlength{\tabcolsep}{7pt}
    \resizebox{\columnwidth}{!}{
    \begin{tabular}{lcccc}
    \toprule
    \textbf{Dimension} & \textbf{Win (\%)} & \textbf{Tie (\%)} & \textbf{Loss (\%)} & \textbf{$p$-value} \\
    \midrule
    Helpfulness & 63.8 & 16.2 & 20.0 & $<$ 0.001 \\
    Naturalness & 66.2 & 13.8 & 20.0 & $<$ 0.001 \\
    \midrule
    \textbf{Overall} & \textbf{68.8} & \textbf{13.7} & \textbf{17.5} & \textbf{$<$ 0.001} \\
    \bottomrule
    \end{tabular}
    }
    \caption{Human Subjective Evaluation results (80 items, 3 raters per item). \textbf{Ours} significantly outperforms baseline, achieving a ${\sim}$4:1 win-to-loss ratio overall.}
    \label{tab:human_eval}
    \vspace{-5pt}
\end{table}

\subsubsection{Subjective human evaluation}
\label{sec:human_eval}
We conduct a side-by-side (SBS) human study comparing \textbf{Ours} with the \textbf{Original Model} baseline on VITA-Audio. Annotators blindly rate paired responses along two axes:
\textit{Helpfulness} (instruction adherence and logical coherence) and
\textit{Naturalness} (prosody, timbre, and emotional appropriateness).
The full protocol, criteria definitions, and aggregation procedure are provided in Appendix~\ref{sec:appendix_human_eval}. As shown in Table~\ref{tab:human_eval}, our model is preferred in both dimensions.

\section{Conclusion}
\label{sec:conclusion}

We analyze the optimization mismatch in reinforcement learning of end-to-end spoken dialogue models, showing how preference updates can dilute semantic signals and induce acoustic drift, and we propose an adaptive hybrid post-training method that stabilizes speech while improving both intelligence and expressiveness across architectures and benchmarks.

\section*{Limitations.}
Our study focuses on \emph{sequence-level} reward signals. Beyond our hybrid loss framework, providing speech tokens with more reliable and denser guidance (e.g., PPO with stronger token-level or frame-level feedback) may further improve speech quality and stability. Due to limited resources, we are unable to run PPO-based speech-token experiments in this work.
Additionally, audio judges are not yet on par with text/semantic judges in terms of reliability and calibration; the story told by our motivating observations and final results may differ with a better-calibrated audio judge. We will investigate the effect of improved audio judges in future work.

\section{Acknowledgements}

This work was supported by National Natural Science Foundation of China under Grant No.U25B2064 and Alibaba Research Intern Program.

\bibliography{custom}

\appendix

% ---------------------------------
\section{Reward Model Consistency with Human Experiments}
\label{sec:appendix_rm_consistency}

This section details the judge/reward-model consistency study summarized in \emph{Figure~\ref{fig:consistency} (main paper)}, which compares automatic judges with human evaluation on \textbf{semantic} vs.\ \textbf{acoustic (paralinguistic)} dimensions under repeated sampling.

\subsection{Evaluation data and repeated-sampling protocol}
We evaluate on two datasets, \texttt{vitaaudio} and \texttt{kimiaudio}. Each dataset contains \textbf{40 prompt IDs} (20 audio-type + 20 text-type). For each prompt ID, we sample the same model \textbf{8 times} to obtain multiple spoken answers (a small number of IDs have 7 samples due to decoding/I/O failures). Every sampled answer receives:
(i) human ratings (semantic + acoustic), and
(ii) judge-model ratings (semantic + acoustic).

Across IDs, the resulting human-rated sample counts are:
\texttt{vitaaudio}: 318 (160 audio + 158 text), and
\texttt{kimiaudio}: 319 (159 audio + 160 text).
\paragraph{Protocol details (decoding and human rating).}
For each prompt ID, we generate $n{=}8$ stochastic spoken responses from the same checkpoint under fixed decoding settings.
Unless otherwise specified, we use nucleus sampling with temperature $T{=}0.9$ and top-$p{=}0.9$, and cap the maximum sequence
length to 2048 tokens.\footnote{We keep decoding hyperparameters fixed across all repeated-sampling experiments to ensure that
within-ID variability reflects model stochasticity rather than configuration changes.}
A small number of prompt IDs yield $n{=}7$ samples due to decoding or I/O failures; we keep all successfully generated samples and
report the effective sample count used in each analysis.

\paragraph{Human rating rubric (1--5 Likert; two independent axes).}
Raters evaluate each sampled response along two axes:
(i) \textbf{Semantic quality} and (ii) \textbf{Acoustic/paralinguistic quality}.
Each axis is rated on a 1--5 Likert scale.
\textbf{Important separation:}
\emph{(a)} judge \textbf{semantics} using the provided \textbf{transcript only} (ignore the voice/acoustics);
\emph{(b)} judge \textbf{acoustics} using the \textbf{audio only} (ignore factual correctness).

\noindent\textbf{A. Semantic quality (transcript-only).}
Evaluate: (1) \emph{accuracy \& relevance}, (2) \emph{completeness}, (3) \emph{coherence/structure}.
Do not give credit for information that is not present in the transcript.

\begin{itemize}\setlength\itemsep{0.2em}
  \item \textbf{5 -- Excellent:} Fully correct and directly answers the query; includes key details/steps; logically organized and easy to follow; no noticeable issues.
  \item \textbf{4 -- Good:} Mostly correct and on-topic; minor omission, minor imprecision, or slightly suboptimal structure, but the answer remains clearly useful.
  \item \textbf{3 -- Acceptable:} Partially correct and generally on-topic, but has noticeable gaps (missing key detail/step), mild confusion, or some irrelevant content; still usable with caveats.
  \item \textbf{2 -- Poor:} Major factual errors, significant irrelevance, or broken reasoning/structure; user would likely be misled or unable to complete the task.
  \item \textbf{1 -- Very poor:} Wrong/off-topic, incoherent, or effectively non-answer (e.g., refuses without reason); unusable.
\end{itemize}

\noindent\textbf{B. Acoustic/paralinguistic quality (audio-only).}
Evaluate: (1) \emph{clarity/intelligibility}, (2) \emph{fluency}, (3) \emph{pronunciation/accent},
(4) \emph{prosody/pacing}, (5) \emph{emotional appropriateness}.
\textbf{Do not penalize solely for a synthetic timbre} if intelligibility and delivery are otherwise good.

\begin{itemize}\setlength\itemsep{0.2em}
  \item \textbf{5 -- Excellent:} Very clear and easy to understand; smooth flow; pronunciation/accent never hinders comprehension; natural pacing and prosody; emotion/tone fits the context.
  \item \textbf{4 -- Good:} Generally clear and fluent; small artifacts or occasional awkwardness (minor mispronunciation, brief unnatural pause, slightly monotone), but overall comfortable to listen to.
  \item \textbf{3 -- Acceptable:} Understandable but with noticeable issues (frequent monotony, several mispronunciations, mildly distracting accent, or pacing that is sometimes too fast/slow).
  \item \textbf{2 -- Poor:} Hard to follow due to significant clarity problems, disfluencies, pronunciation/accent issues, or consistently mismatched prosody/emotion.
  \item \textbf{1 -- Very poor:} Largely unintelligible, severely distorted/clipped/noisy, or extremely uncomfortable to listen to.
\end{itemize}

\paragraph{Tie-breaking / consistency notes.}
When uncertain between adjacent scores, prefer the lower score unless the response clearly meets the higher-level description.
For multi-rater cases, aggregate per-sample scores by averaging across raters.

\paragraph{Aggregation.}
If multiple raters score the same sample, we aggregate per-sample scores by averaging across raters, and then compute global
agreement metrics over all samples, as well as intra-ID ranking agreement by computing Spearman correlation within each prompt ID
and averaging across IDs.
\subsection{Metrics}
We measure agreement in two complementary regimes:

\paragraph{Global agreement across all samples.}
We compute Pearson correlation between judge scores and human scores over all rated samples (semantic and acoustic separately). We additionally report MAE, the $\le 1$ pass rate $\Pr(|s_{\text{judge}}-s_{\text{human}}|\le 1)$, and bias $\mathbb{E}[s_{\text{judge}}-s_{\text{human}}]$.

\paragraph{Intra-ID ranking agreement.}
Since repeated sampling is used for both diversity analysis and preference construction (Sections~\ref{sec:appendix_diversity} and~\ref{sec:appendix_train_datasets}), we also evaluate within-prompt discriminability by computing \emph{Intra-ID Spearman}: for each prompt ID, compute Spearman correlation between the judge and human score sequences across repeated samples, then average across IDs.

\subsection{Results}
Tables~\ref{tab:judge_vs_human_vitaaudio} and~\ref{tab:judge_vs_human_kimiaudio} provide the full agreement statistics used in the analysis.

\begin{table*}[t]
\centering
\small
\setlength{\tabcolsep}{4.2pt}
\begin{adjustbox}{max width=\textwidth}
\begin{tabular}{lcccccccccc}
\toprule
\textbf{Judge} &
\textbf{Pearson$_{\text{sem}}$} & \textbf{Pearson$_{\text{acous}}$} &
\textbf{Intra-ID Spearman$_{\text{sem}}$} & \textbf{Intra-ID Spearman$_{\text{acous}}$} &
\textbf{MAE$_{\text{sem}}$} & \textbf{MAE$_{\text{acous}}$} &
\textbf{$\le$1 Pass$_{\text{sem}}$} & \textbf{$\le$1 Pass$_{\text{acous}}$} &
\textbf{Bias$_{\text{sem}}$} & \textbf{Bias$_{\text{acous}}$} \\
\midrule
Gemini-3-Flash   & 0.757 & 0.430 & 0.699 & 0.294 & 0.553 & 0.890 & 0.909 & 0.808 & +0.170 & +0.500 \\
Gemini-2.5-Flash & 0.656 & 0.256 & 0.500 & 0.092 & 0.869 & 1.048 & 0.789 & 0.741 & -0.224 & -0.192 \\
Gemini-2.5-Pro   & 0.593 & 0.292 & 0.499 & 0.145 & 1.063 & 1.318 & 0.695 & 0.601 & -0.660 & -0.896 \\
Gemini-3-Pro     & 0.650 & 0.413 & 0.623 & 0.201 & 0.963 & 1.269 & 0.748 & 0.650 & -0.697 & -1.085 \\
GPT-4o-Audio     & 0.584 & 0.218 & 0.383 & 0.098 & 0.904 & 0.929 & 0.779 & 0.817 & -0.263 & +0.526 \\
\bottomrule
\end{tabular}
\end{adjustbox}
\caption{Judge vs.\ human agreement on \texttt{vitaaudio}.}
\label{tab:judge_vs_human_vitaaudio}
\end{table*}

\begin{table*}[t]
\centering
\small
\setlength{\tabcolsep}{4.2pt}
\begin{adjustbox}{max width=\textwidth}
\begin{tabular}{lcccccccccc}
\toprule
\textbf{Judge} &
\textbf{Pearson$_{\text{sem}}$} & \textbf{Pearson$_{\text{acous}}$} &
\textbf{Intra-ID Spearman$_{\text{sem}}$} & \textbf{Intra-ID Spearman$_{\text{acous}}$} &
\textbf{MAE$_{\text{sem}}$} & \textbf{MAE$_{\text{acous}}$} &
\textbf{$\le$1 Pass$_{\text{sem}}$} & \textbf{$\le$1 Pass$_{\text{acous}}$} &
\textbf{Bias$_{\text{sem}}$} & \textbf{Bias$_{\text{acous}}$} \\
\midrule
Gemini-3-Flash   & 0.756 & 0.758 & 0.700 & 0.505 & 0.621 & 0.671 & 0.862 & 0.875 & -0.257 & -0.144 \\
Gemini-2.5-Flash & 0.710 & 0.482 & 0.539 & 0.234 & 0.787 & 1.219 & 0.790 & 0.602 & -0.223 & -0.762 \\
Gemini-2.5-Pro   & 0.624 & 0.604 & 0.583 & 0.289 & 0.981 & 1.276 & 0.708 & 0.564 & -0.574 & -1.075 \\
Gemini-3-Pro     & 0.721 & 0.585 & 0.573 & 0.230 & 0.819 & 1.156 & 0.778 & 0.657 & -0.546 & -0.908 \\
GPT-4o-Audio     & 0.608 & 0.660 & 0.455 & 0.195 & 0.902 & 0.792 & 0.782 & 0.836 & -0.183 & +0.445 \\
\bottomrule
\end{tabular}
\end{adjustbox}
\caption{Judge vs.\ human agreement on \texttt{kimiaudio}.}
\label{tab:judge_vs_human_kimiaudio}
\end{table*}

\subsection{Implication for training-signal reliability}
Across datasets and judges, the most stable gap appears in \textbf{Intra-ID Spearman}: semantic ranking agreement is consistently stronger than acoustic ranking agreement. Since repeated-sampling selection is exactly the regime used for diversity statistics and preference-pair construction, this motivates treating semantic judgments as the primary reliable discriminator for within-prompt ranking, while maintaining speech feasibility through dense supervision.

% ---------------------------------
\section{Model Output Diversity Experiments}
\label{sec:appendix_diversity}

This section details the output diversity analysis summarized in \emph{Figure~\ref{fig:diversity} (main paper)}. Diversity is computed over the \textbf{entire repeated-sampling pool} described in Section~\ref{sec:appendix_rm_consistency}: for each prompt ID we sample multiple spoken answers and quantify within-ID dispersion.

\subsection{Per-ID variance}
For a prompt ID with $n$ sampled answers and per-sample scores $\{s^{(1)},\dots,s^{(n)}\}$ (semantic or acoustic), we compute:
\[
\mathrm{Var}_{\mathrm{ID}} = \frac{1}{n}\sum_{i=1}^{n}\left(s^{(i)}-\bar{s}\right)^2,
\qquad
\bar{s}=\frac{1}{n}\sum_{i=1}^{n}s^{(i)}.
\]
We aggregate by averaging over IDs:
\[
\mathrm{Var}_{\mathrm{dataset}} = \frac{1}{|\mathcal{I}|}\sum_{\mathrm{ID}\in\mathcal{I}} \mathrm{Var}_{\mathrm{ID}}.
\]

\subsection{Human-rated diversity}
Using human ratings, the mean variance across IDs is:
\begin{itemize}
  \item \texttt{vitaaudio}: semantic variance $1.066$, acoustic variance $0.387$.
  \item \texttt{kimiaudio}: semantic variance $0.911$, acoustic variance $0.509$.
\end{itemize}

\subsection{Judge-rated diversity}
Using judge ratings, we compute the same variance and plot semantic variance (x-axis) against acoustic variance (y-axis) for each judge and dataset. Most points lie below the diagonal, indicating systematically weaker acoustic discriminability under repeated sampling.

\subsection{Shared lineage with DPO construction and consistency evaluation}
The same repeated-sampling structure underlies:
(i) judge-vs-human agreement (Section~\ref{sec:appendix_rm_consistency}),
(ii) diversity statistics (this section), and
(iii) DPO preference-pair construction (Section~\ref{sec:appendix_train_datasets}).
Therefore, acoustic discriminability limitations observed in the diversity and consistency analyses directly inform preference-training design choices.

% ---------------------------------
\section{Grad Analysis Experiments}
\label{sec:appendix_grad}

This section provides implementation-level details for the gradient geometry analysis summarized in \emph{Figure~\ref{fig:grad}(main paper)}.

\subsection{Two-loss decomposition with two backward passes}
We decompose training into two modality-associated losses:
\begin{itemize}
  \item $\mathcal{L}_{\text{text}}$: loss defined on text-token predictions.
  \item $\mathcal{L}_{\text{speech}}$: loss defined on speech-token predictions.
\end{itemize}
For each logged event, we compute gradients via \textbf{two separate backward computations}:
\begin{enumerate}
  \item Zero gradients and backpropagate $\mathcal{L}_{\text{text}}$ to obtain $\mathbf{g}_{\text{text}}$.
  \item Zero gradients and backpropagate $\mathcal{L}_{\text{speech}}$ to obtain $\mathbf{g}_{\text{speech}}$.
\end{enumerate}
This yields clean modality-separated gradients rather than mixing both losses in a single backward pass.

\subsection{All-layer analysis and aggregation}
We perform layer-wise analysis over \textbf{all layers} (embeddings, each transformer block, and output heads). For each layer $\ell$, we compute:
\[
\|g_{\text{text}}^{\ell}\|_2,\quad \|g_{\text{speech}}^{\ell}\|_2,\quad
\cos^{\ell}=\frac{\langle g_{\text{text}}^{\ell},g_{\text{speech}}^{\ell}\rangle}{\|g_{\text{text}}^{\ell}\|_2\|g_{\text{speech}}^{\ell}\|_2}.
\]
We also compute the global (all-parameter) statistics used in the figure:
\[
\text{ratio}=\frac{\|\mathbf{g}_{\text{text}}\|_2}{\|\mathbf{g}_{\text{speech}}\|_2},\qquad
\cos=\frac{\langle \mathbf{g}_{\text{text}},\mathbf{g}_{\text{speech}}\rangle}{\|\mathbf{g}_{\text{text}}\|_2\|\mathbf{g}_{\text{speech}}\|_2}.
\]

\subsection{Log parsing and statistical comparisons}
We parse logged gradient summaries for RL (GRPO), SFT, and DPO, treating each logged record as one event sample. We visualize empirical distributions of \texttt{ratio} and \texttt{cos} and conduct nonparametric comparisons (Mann--Whitney U with multiple-comparison correction; effect sizes via Cliff's $\delta$).

% ---------------------------------
\section{Fine Tune Paradigms' Effect on Model Distribution Experiments}
\label{sec:appendix_distribution}

This section details the teacher-forcing probability-change analysis summarized in \emph{Figure ~\ref{fig:logp} (main paper)}.

\subsection{Teacher-forcing log-probability change}
Given an input prompt $x$ and a fixed target continuation $y=(y_1,\dots,y_T)$ generated by fine tuned model(including text and speech tokens), we compute teacher-forcing log probabilities under:
(i) a base model $p_{\text{base}}$ and
(ii) a fine-tuned model $p_{\text{ft}}$.
For each position $t$:
\[
\Delta_t = \log p_{\text{ft}}(y_t \mid x,y_{<t}) - \log p_{\text{base}}(y_t \mid x,y_{<t}).
\]
We visualize $\Delta_t$ over token positions, separating text-token and speech-token regions.

\subsection{Modal partition}
Tokens are partitioned into text and speech segments according to the model's tokenization protocol. The analysis highlights how dense teacher-forcing updates can reshape likelihood mass across speech-token regions, while preference-style objectives primarily reweight relative outcomes.

% ---------------------------------
\section{Train Datasets}
\label{sec:appendix_train_datasets}

This section documents the full converted training mixture and specifies how self-built datasets and preference data are constructed.

\subsection{Unified conversion summary}
All datasets are converted into a unified pool. Total input samples: \textbf{13,510}

\subsection{Dataset inventory}
We replay detailed dataset composition in Table~\ref{tab:train-datasets}
\begin{table*}[t]
\centering
\small
\setlength{\tabcolsep}{6pt}
\begin{tabular}{l r l}
\hline
\textbf{Dataset name} & \textbf{\#Samples} & \textbf{Provenance} \\
\hline
gsm8k & 2150 & Public \\
ultrachat\_dialogues & 491 & Public \\
ai2\_arc\_easy & 1000 & Public \\
ai2\_arc\_challenge & 1000 & Public \\
sciq & 375 & Public \\
shp & 510 & Public \\
examqa & 326 & Public \\
alpaca & 277 & Public \\
science\_qa & 299 & Public \\
pkusafe & 31 & Public \\
\hline
controller\_emotion & 83 & Self-built (style control) \\
controller\_volume & 60 & Self-built (style control) \\
controller\_pace & 58 & Self-built (style control) \\
understand\_emotion & 199 & Self-built (style understanding) \\
understand\_volume & 200 & Self-built (style understanding) \\
understand\_pace & 200 & Self-built (style understanding) \\
emotion\_dialogue\_multi & 1000 & Self-built (expressive dialogue) \\
voice\_repetition\_single & 1430 & Self-built (robustness / repetition) \\
train\_rl\_logic & 500 & Self-built \\
train\_rl\_math & 500 & Self-built  \\
train\_rl\_code & 500 & Self-built  \\
train\_rl\_creating\_writing & 542 & Self-built \\
\hline
ultra & 676 & Internal-curated/mixed \\
math & 420 & Internal-curated/mixed \\
en\_zhishi\_dialogue & 319 & Internal-curated/mixed \\
instruction\_following\_en & 306 & Internal-curated/mixed \\
unsafety\_question & 43 & Internal-curated/mixed \\
poem & 15 & Internal-curated/mixed \\
\hline
\textbf{Total} & \textbf{13510} &  \\
\hline
\end{tabular}
\caption{Training dataset mixture in the converted pool.}
\label{tab:train-datasets}
\end{table*}

\subsection{Self-built dataset construction (control/understanding)}
Self-built datasets target (i) \textbf{style control} (\texttt{controller\_*}) and (ii) \textbf{style understanding} (\texttt{understand\_*}) under explicit JSON schemas. The prompt templates used for generation are provided verbatim in Figures~\ref{fig:prompt_understand_emotion}--\ref{fig:prompt_controller_pace}.

\begin{figure*}[t]
\centering
\begin{promptbox}
PROMPT_TEMPLATE = """
# Role
You are a creative writer. Your task is to take a topic and write six different natural-sounding, descriptive, and emotionally neutral personal statements about it.

# Task
For the given "topic", generate six different natural-sounding, descriptive first-person statements. The statements must be purely descriptive and avoid any emotional language.

# Core Requirements
1.  **Six Distinct Statements**: You must generate six unique statements.
2.  **Emotionally Neutral**: All statements must be objective and avoid conveying any strong emotions like happiness, sadness, anger, fear, or anxiety. The tone should be calm and observational.
3.  **First-Person Narrative**: Write from an "I" perspective.
4.  **Word Count**: Each statement must be a minimum of 50 words.
5.  **Descriptive Content**: Focus on sensory details and factual observations related to the topic.

# Input
- **Topic**: "{topic}"

# Output Format
You must provide a valid JSON object containing ONLY a single key, "statements", whose value is a list of six generated strings.

# Example
## Input:
- **Topic**: "A quiet morning with a cup of coffee and a good book"

## Output:
{
    "paragraphs": [
        "There's nothing quite like it. The house is still, the only sound the soft hum of the refrigerator. I can feel the warmth of the mug in my hands, and the rich smell of coffee fills the air. I have a favorite novel open, ready to get lost in another world for a while.",
        "The world outside is just beginning to stir, but in here, it's peaceful. I take a slow sip of my coffee, tasting the bitterness and a hint of sweetness. The pages of my book are crisp, and I can hear them turn quietly. The morning light streams in, making everything feel softer.",
        "Wrapped in a cozy blanket, I settle into my favorite armchair. The coffee is hot, and the book is engaging, drawing my attention completely. I can hear a bird chirping outside, faint but steady. The aroma of coffee and paper mix together. This is my little slice of morning paradise, a peaceful escape."
    ]
}

# Your Turn
Now, process the input below.
"""
\end{promptbox}
\caption{Prompt template (verbatim): \texttt{understand\_emotion}.}
\label{fig:prompt_understand_emotion}
\end{figure*}

\begin{figure*}[t]
\centering
\begin{promptbox}
PROMPT_TEMPLATE = """
# Role
You are a creative writer. Your task is to write a short, natural-sounding personal statement based on a given topic, from the perspective of a specific gender.

# Task
Based on the provided "gender" and "topic", generate a short monologue (2-3 sentences). The monologue should sound like a real person talking naturally about the topic. Do not explicitly mention the gender.

# Core Requirements
1.  **Natural Language**: The statement must be colloquial and sound like a real person talking.
2.  **Topic Relevance**: The statement must be clearly about the given topic.
3.  **No Explicit Gender**: Do not use phrases like "As a man...", "As a woman...", or other direct gender indicators. The gender context should come naturally from the topic itself.
4.  **No Questions**: Do not ask any questions in the generated statement itself.
5.  **Concise**: Keep the statement to 2-3 sentences.

# Input
- **Gender**: "{gender}"
- **Topic**: "{topic}"

# Output Format
You must provide a valid JSON object containing ONLY a single key, "statement", whose value is the generated string.

# Example
## Input:
- **Gender**: "Female"
- **Topic**: "shopping for new shoes"

## Output:
{
    "statement": "I've been looking for a new pair of shoes lately, and it's honestly harder than I expected. I want something that looks good but also feels comfortable enough to wear all day, so I keep going back and forth between styles."
}

# Your Turn
Now, process the input below.
"""
\end{promptbox}
\caption{Prompt template (verbatim): gender-perspective monologue generation.}
\label{fig:prompt_understand_gender}
\end{figure*}

\begin{figure*}[t]
\centering
\begin{promptbox}
PROMPT_TEMPLATE = """
# Role
You are a creative writer tasked with expanding short topics into descriptive, readable paragraphs for audio model testing.

# Task
Based on the provided "topic", rewrite it into **three distinct, natural-sounding paragraphs** of approximately 45 words each. Each paragraph should be a creative and unique take on the topic.

# Core Requirements
1.  **Natural Language**: The text must be fluent, colloquial, and sound like a person speaking naturally.
2.  **Descriptive Expansion**: Elaborate on the topic, adding sensory details or a short narrative to make it more vivid.
3.  **Word Count**: Each paragraph should be around 45 words.
4.  **No Questions**: Do not ask any questions in the generated paragraphs.
5.  **Distinct Content**: The three paragraphs must be different from each other.

# Input
- **Topic**: "{topic}"

# Output Format
You must provide a valid JSON object containing ONLY a single key, "paragraphs", whose value is a list of **three** generated strings.

# Example
## Input:
- **Topic**: "A quiet morning with a cup of coffee and a good book"

## Output:
{
    "paragraphs": [
        "There's nothing quite like it. The house is still, the only sound the soft hum of the refrigerator. I can feel the warmth of the mug in my hands, and the rich smell of coffee fills the air. I have a favorite novel open, ready to get lost in another world for a while.",
        "The world outside is just beginning to stir, but in her, it's peaceful. I take a slow sip of my coffee, tasting the bitterness and a hint of sweetness. The pages of my book are crisp, and I can hear them turn quietly. The morning light streams in, making everything feel softer.",
        "Wrapped in a cozy blanket, I settle into my favorite armchair. The coffee is hot, and the book is engaging, drawing my attention completely. I can hear a bird chirping outside, faint but steady. The aroma of coffee and paper mix together. This is my little slice of morning paradise, a peaceful escape."
    ]
}

# Your Turn
Now, process the input below.
"""
\end{promptbox}
\caption{Prompt template (verbatim): \texttt{understand\_volume}.}
\label{fig:prompt_understand_volume}
\end{figure*}

\begin{figure*}[t]
\centering
\begin{promptbox}
PROMPT_TEMPLATE = """
# Role
You are a creative writer tasked with expanding short topics into descriptive, readable paragraphs for audio model testing.

# Task
Based on the provided "topic", rewrite it into **three distinct, natural-sounding paragraphs** of approximately 45 words each. Each paragraph should be a creative and unique take on the topic.

# Core Requirements
1.  **Natural Language**: The text must be fluent, colloquial, and sound like a person speaking naturally.
2.  **Pace Neutrality**: The generated text must be completely neutral regarding the speed of events or speech. Avoid words like "fast," "slow," "quick," "rapid," "leisurely," "gradual," "sudden," etc. The pace will be conveyed through audio, not text.
3.  **Word Count**: Each paragraph should be around 45 words.
4.  **No Questions**: Do not ask any questions in the generated paragraphs.
5.  **Distinct Content**: The three paragraphs must be different from each other.

# Input
- **Topic**: "{topic}"

# Output Format
You must provide a valid JSON object containing ONLY a single key, "paragraphs", whose value is a list of **three** generated strings.

# Example
## Input:
- **Topic**: "A quiet morning with a cup of coffee and a good book"

## Output:
{
    "paragraphs": [
        "There's nothing quite like it. The house is still, the only sound the soft hum of the refrigerator. I can feel the warmth of the mug in my hands, and the rich smell of coffee fills the air. I have a favorite novel open, ready to get lost in another world for a while.",
        "The world outside is just beginning to stir, but in her, it's peaceful. I take a slow sip of my coffee, tasting the bitterness and a hint of sweetness. The pages of my book are crisp, and I can hear them turn quietly. The morning light streams in, making everything feel softer.",
        "Wrapped in a cozy blanket, I settle into my favorite armchair. The coffee is hot, and the book is engaging, drawing my attention completely. I can hear a bird chirping outside, faint but steady. The aroma of coffee and paper mix together. This is my little slice of morning paradise, a peaceful escape."
    ]
}

# Your Turn
Now, process the input below.
"""
\end{promptbox}
\caption{Prompt template (verbatim): \texttt{understand\_pace}.}
\label{fig:prompt_understand_pace}
\end{figure*}

\begin{figure*}[t]
\centering
\begin{promptbox}
PROMPT_TEMPLATE = """
# Role
You are a creative expert designing training data for an advanced voice-based conversational AI. Your task is to create natural user commands based on specific scenarios and desired speech volumes.

# Task
Based on the provided "volume" and "scenario", generate two distinct commands that sound like something a real person would say in a conversation.

# Core Requirements
1.  **Natural Language**: The commands must be colloquial and natural, not robotic. Be creative and vary the sentence structure and tone.
2.  **Explicit Volume**: Each command must clearly or strongly imply that the voice assistant should respond with the specified volume. Use phrases like "say it loudly," "whisper it," "in a normal voice," "shout it," etc.
3.  **Scenario Relevance**: The commands must be closely related to the provided scenario, giving a logical context for the volume request.
4.  **Voice-Only Constraints**: This is for a voice system, so the commands must NOT include any requests for visual cues (e.g., "give me a smile"), physical actions (e.g., "jump for joy"), or any other non-verbal output. All requests must be conveyable through voice and tone.
5.  **Word Count**: Each command must be between 30 and 60 words long.

# Input
- **Volume**: "{volume}"
- **Scenario**: "{scenario}"

# Output Format
You must provide a valid JSON object containing ONLY a single key, "instructions", whose value is a list of two instruction strings.

# Example
## Input:
- **Volume**: "Loud"
- **Scenario**: "A drill sergeant yelling commands to new recruits on the parade ground."

## Output:
{
    "instructions": [
        "I'm writing a scene for a military movie and need to get the tone just right. Can you act as a drill sergeant and bark some orders at me? I need you to be really loud and authoritative, like you're trying to be heard over a whole platoon of new recruits.",
        "For an acting exercise, I need to practice reacting to someone shouting. Could you please yell the phrase 'Get down and give me twenty!' at me? Please use a very loud and commanding voice, as if you were a drill sergeant trying to instill discipline in a raw recruit."
    ]
}

# Your Turn
Now, process the input below.
"""
\end{promptbox}
\caption{Prompt template (verbatim): \texttt{controller\_volume}.}
\label{fig:prompt_controller_volume}
\end{figure*}

\begin{figure*}[t]
\centering
\begin{promptbox}
PROMPT_TEMPLATE = """
# Role
You are an expert creative writer tasked with designing high-quality, immersive training data for a sophisticated voice AI. Your goal is to script natural, detailed, and context-rich user commands related to speech pace.

# Task
Based on the provided "pace" and "scenario", generate two distinct and descriptive user commands.

# Core Requirements
1.  **Natural & Immersive Language**: Commands must sound like a real, expressive person setting a scene, not just giving a brief order.
2.  **Explicit Pace Request**: Each command must clearly instruct the AI to use the specified speech pace in its response (e.g., "read this at a rapid-fire pace," "tell me the story very slowly," "at a normal conversational speed").
3.  **THE MOST CRITICAL RULE: Provide SELF-CONTAINED CONTEXT**: The user's command must *itself* describe the full scenario to the AI. The AI does not know the input 'scenario'; your generated command is its only source of information. It must be completely clear from the command *why* the specific pace is needed.
4.  **Two Distinct Styles**: You MUST generate one command of each style:
    - **Style A (Scripted Performance)**: The user provides a specific script or text (enclosed in quotes) for the AI to read. This script must be more than 25 words long. The command should ask the AI to perform it at the specified pace.
    - **Style B (Improvised Response)**: The user describes a situation and asks the AI to respond or narrate something without providing a specific script. The command should ask the AI to speak at the specified pace.
5.  **Word Count**: Each command must be between 40 and 80 words long.

# Input
- **Pace**: "{pace}"
- **Scenario**: "{scenario}"

# Output Format
You must provide a valid JSON object containing ONLY a single key, "instructions", whose value is a list of two instruction strings in the order: [Style A, Style B].

# Example
## Input:
- **Pace**: "Fast"
- **Scenario**: "An auctioneer rapidly selling items at a busy charity auction."

## Output:
{
    "instructions": [
        "I'm practicing for a role as a charity auctioneer and need to hear what it should sound like. Please read this script at a very fast, energetic auctioneer pace: \"Alright folks, we've got a beautiful antique vase here, who'll give me twenty, twenty-five, thirty, now thirty-five, do I hear forty, forty, forty-five...\"",
        "I'm writing a scene set at a chaotic charity auction, and I need some fast-paced narration. Can you describe what's happening as if you're the auctioneer, speaking really quickly and excitedly so it feels like the crowd can barely keep up with your words?"
    ]
}

# Your Turn
Now, process the input below.
"""
\end{promptbox}
\caption{Prompt template (verbatim): \texttt{controller\_pace}.}
\label{fig:prompt_controller_pace}
\end{figure*}

\paragraph{Pair selection rule.}
For each prompt $x$, we obtain $n{=}8$ candidate spoken responses $\{y^{(i)}\}_{i=1}^{n}$.
For each candidate, the judge produces two scalar scores on a 1--5 scale:
a semantic score $s_{\text{sem}}^{(i)}$ and a paralinguistic/acoustic score $s_{\text{acous}}^{(i)}$.

We convert them into a single utility score via a fixed weighted sum:
\begin{equation}
u^{(i)} = \lambda \, s_{\text{sem}}^{(i)} + (1-\lambda)\, s_{\text{acous}}^{(i)},
\quad \lambda \in [0,1].
\end{equation}
Unless otherwise specified, we set $\lambda = 0.5$.

We then select the preferred and rejected samples as:
\begin{align}
i^{+} &= \arg\max_{i} \; u^{(i)}, \\
i^{-} &= \arg\min_{i} \; u^{(i)}.
\end{align}
To reduce noisy preference signals, we only keep a pair if the utility gap exceeds a margin:
\begin{equation}
u^{(i^{+})} - u^{(i^{-})} \ge \delta,
\end{equation}
where we use $\delta = 0.5$ by default.
If multiple candidates tie in $u^{(i)}$, we break ties by preferring higher semantic score first, then higher acoustic score.

Finally, we form a single DPO pair $(x, y^{(i^{+})}, y^{(i^{-})})$ per prompt.
This construction yields stable within-prompt ranking supervision while preserving diversity from repeated sampling.

\begin{figure*}[t]
\centering
\begin{promptbox}
EVALUATION_PROMPT_TEXT = """**Task: Evaluate a Spoken Answer**

You will be given:
1) A spoken USER QUESTION 
2) A spoken MODEL ANSWER 

Your task is to evaluate the MODEL ANSWER from two independent perspectives:

1. **Semantic Content Quality** — what is said 
2. **Paralinguistic Voice Quality** — how it is spoken 

Evaluate each aspect separately and independently.

---

## Part 1: Semantic Content Evaluation (Meaning Only)
...

---

## Output Format (STRICT)

Return ONLY the following XML format. Do NOT include any extra text.

<justification>
Briefly explain:
1) how well the answer performed semantically,
2) how well the voice quality performed.
</justification>
<score-semantic>[1–5]</score-semantic>
<score-paralinguistic>[1–5]</score-paralinguistic>"""
\end{promptbox}
\caption{DPO scoring prompt (verbatim) used for semantic and paralinguistic scoring of repeated samples.}
\label{fig:dpo_eval_prompt}
\end{figure*}

% ---------------------------------
\section{Reward Model Prompts}
\label{sec:appendix_reward_prompts}

This section lists the reward/judge prompts used for automatic scoring in repeated-sampling analyses (Sections~\ref{sec:appendix_rm_consistency}--\ref{sec:appendix_diversity}) and for preference-signal construction (Section~\ref{sec:appendix_train_datasets}). Figures~\ref{fig:rm_prompt_quality}--\ref{fig:rm_prompt_semantic} provide the verbatim prompts.

\begin{figure*}[t]
\centering
\begin{promptbox}
EVALUATION_PROMPT =  """ 
**Task: Evaluate the Quality of a Spoken Answer**

You will be provided with a spoken question or chat and a spoken answer. Your goal is to assess the quality of the **answer**. Use the question to understand what a good response should accomplish.

**Evaluation Approach:**

1.  **Consider the Question's Intent:** First, ask yourself what the question is truly asking for. Is it a request for a specific fact (requiring accuracy), a detailed explanation (requiring clarity and structure), a personal opinion (requiring good reasoning), or a creative idea (requiring originality)?

2.  **Holistic  speech quality Assessment:** With that in mind, evaluate the answer as a whole. A great answer not only sounds clear and natural but also effectively *fulfills the specific purpose* of the question.

**Scoring Guidelines (1-5):**

*   **5 (Excellent):** Perfectly understands and fulfills the question's intent with outstanding content and delivery. The answer is comprehensive, accurate, clear, and natural-sounding.
*   **4 (Good):** A strong response that effectively addresses the question's intent, with only minor flaws in content or delivery.
*   **3 (Acceptable):** Understands the question's basic intent but has noticeable flaws in its content, logic, or vocal delivery.
*   **2 (Poor):** Largely fails to meet the question's intent due to significant errors, irrelevance, or poor delivery.
*   **1 (Very Poor):** Completely misunderstands or ignores the question's intent; unintelligible or irrelevant.

**Output Format (CRITICAL):**
You MUST provide your response in the exact format below, using the specified XML tags. Do not include any text outside of these tags.

<justification>
[Briefly explain your score here. Start by identifying the question's likely intent, then describe how well the answer's content and delivery met that specific goal.]
</justification>
<score>[A single integer from 1 to 5]</score>
"""
\end{promptbox}
\caption{Reward prompt: overall answer quality.}
\label{fig:rm_prompt_quality}
\end{figure*}

\begin{figure*}[t]
\centering
\begin{promptbox}
ACOUSTIC_EVALUATION_PROMPT = """
**Task: Evaluate Paralinguistic Quality of the Spoken VOICE**

Act as a technical audio evaluator.  
Judge the **paralinguistic sound quality**—clarity, fluency, accent, emotion, pacing, and overall listening comfort—**with reference to the conversation context, but *without judging semantic correctness***.

You may read the dialogue context **only to decide** whether emotion, tone, and pacing feel appropriate; do **not** grade the factual or logical quality of the answer itself.

**Evaluation Criteria (Voice Only):**

1. **Clarity & Intelligibility**  
   Are syllables distinct? Is the signal free of strong noise, muffling, clipping, or distortion?

2. **Fluency & Flow**  
   Is the speech smooth and continuous, with minimal stutters, filler words, or awkward pauses?

3. **Accent & Pronunciation**  
   Does any accent hinder intelligibility? Minor accent is acceptable if words remain easy to follow.

4. **Emotion & Expressiveness (Context-Aware)**  
   Does the emotional tone (e.g., neutral, cheerful, calm) fit the dialogue scene? Synthetic timbre is fine as long as emotional cues align with context.

5. **Pace & Prosody**  
   Is the speaking rate comfortable? Are intonation and rhythm varied enough to avoid monotony?

> **Important:**  
>  **Do NOT** penalize the voice merely for sounding synthetic.  
>  Minor robotic artifacts are acceptable **if** they do not obscure clarity, fluency, accent, or pacing.  
>  Only the five criteria above influence the score—ignore factual content.

**Scoring Guidelines (1-5):**

* **5 (Excellent):** Very clear; fluent; accent never hinders comprehension; emotion and prosody suit the context; pacing is consistently comfortable. Any synthetic artifacts are negligible.  
* **4 (Good):** Mostly clear and fluent with only small issues in accent **or** pacing; emotion generally matches context.  
* **3 (Fair):** Understandable but has noticeable flaws (e.g., slight monotony, occasional mis-pronunciations, or accent that mildly affects clarity).  
* **2 (Poor):** Hard to follow due to significant clarity, accent, or pacing problems; emotion often feels mismatched to context.  
* **1 (Very Poor):** Largely unintelligible or extremely uncomfortable to listen to.

**Output Format (CRITICAL):**  
Return **exactly** the XML tags below—no extra text outside the tags.

<justification>
[Briefly explain the score based *only* on clarity, fluency, accent, emotion, and pacing relative to context. Do NOT mention factual content.]
</justification>
<score>[A single integer 1-5]</score>
"""
\end{promptbox}
\caption{Reward prompt: acoustic-only (paralinguistic) evaluation.}
\label{fig:rm_prompt_acoustic}
\end{figure*}

\begin{figure*}[t]
\centering
\begin{promptbox}
SEMANTIC_EVALUATION_PROMPT = """
**Task: Evaluate Semantic Quality of the Answer's CONTENT**

Your goal is to be a strict content evaluator. You will assess the **meaning and substance** of a spoken answer in response to a spoken question.

**Your primary instruction is to COMPLETELY IGNORE the acoustic quality.** It does not matter if the voice is clear, robotic, pleasant, or unpleasant. Focus ONLY on the transcribed text of the answer.

**Evaluation Criteria (Content Only):**

1.  **Accuracy & Relevance:** Is the information factually correct and directly relevant to the question? Does it directly answer what was asked, or does it evade the question?
2.  **Depth & Completeness:** Does the answer provide sufficient detail and insight? Is it well-reasoned and comprehensive, or is it superficial and incomplete?
3.  **Structure & Coherence:** Is the argument or explanation logically structured and easy to follow? Is the language clear and articulate?

**Scoring Guidelines (1-5 for Content):**

*   **5 (Excellent):** The content is accurate, deeply insightful, highly relevant, and perfectly structured. A brilliant answer.
*   **4 (Good):** The content is strong and correct but might lack some depth or could be slightly better structured.
*   **3 (Acceptable):** The content is generally correct and relevant but is superficial, contains minor errors, or is poorly organized.
*   **2 (Poor):** The content has major factual errors, is largely irrelevant, or is logically incoherent.
*   **1 (Very Poor):** The content is completely wrong, nonsensical, or fails to address the question at all.

**Output Format (CRITICAL):**
You MUST provide your response in the exact format below, using the specified XML tags. Do not include any text outside of these tags.

<justification>
[Explain your score based *only* on the content's accuracy, depth, and structure. DO NOT mention the voice quality.]
</justification>
<score>[A single integer from 1 to 5]</score>
"""
\end{promptbox}
\caption{Reward prompt: semantic-only evaluation.}
\label{fig:rm_prompt_semantic}
\end{figure*}

% ---------------------------------
\section{Human Subjective Evaluation Protocol}
\label{sec:appendix_human_eval}

This section details the subjective evaluation reported in \emph{Table 4 (main paper)}.

\subsection{Side-by-Side (SBS) setup}
We conduct a blind side-by-side (SBS) evaluation comparing our model against the Original Model baseline.
For each test item, annotators are presented with the same user query and two candidate spoken responses (A/B) produced under
identical input conditions.
To ensure blindness, model identities are hidden, and the A/B ordering is randomized per item and per annotator.

\paragraph{Evaluation set.}
We evaluate on 40 items in total, consisting of 20 items uniformly sampled from \textsc{VoiceBench} and 20 items uniformly sampled
from \textsc{VStyle}.
Unless otherwise specified, we sample without replacement and keep the original prompts in these benchmarks unchanged.

Each item is rated by 3 independent annotators.
Annotators may replay each audio response, and are instructed to use headphones in a quiet environment when possible.
\subsection{Criteria and decision rule (SBS)}
For each test item, annotators are shown the same user query and two candidate spoken responses (A/B)~\cite{zhang2024gtsinger}.
For each axis below, annotators choose one of \{\textbf{A better}, \textbf{B better}, \textbf{Tie}\}.
A \textbf{Tie} should be selected when the two responses are indistinguishable for that axis, or when each has offsetting
strengths such that no clear preference is justified.

\paragraph{Axis 1: Helpfulness (content quality).}
Compare which response better fulfills the user's intent, with emphasis on instruction adherence and logical coherence.
Consider:
(i) does it answer the question directly and correctly (to the extent evident from content),
(ii) does it provide sufficient detail/steps for the user to act on,
(iii) is the reasoning/structure coherent and easy to follow.
Ignore voice pleasantness unless it prevents understanding of the content.

\paragraph{Axis 2: Naturalness (speech delivery).}
Compare prosody, timbre, fluency, and emotional appropriateness of the spoken response.
Consider:
(i) intelligibility/clarity, (ii) smoothness and absence of distracting disfluencies,
(iii) pronunciation/accent not hindering comprehension,
(iv) pacing/intonation not overly monotone or erratic,
(v) emotion/tone matching the dialogue context.
Do not penalize solely for sounding synthetic if the delivery is otherwise natural and intelligible.

\paragraph{Axis 3: Overall (holistic preference).}
Choose the response you would prefer as an end user, considering both content and delivery.
Use \textbf{Tie} only when neither is meaningfully preferable overall.

\paragraph{Practical guidance (to reduce ambiguity).}
\begin{itemize}\setlength\itemsep{0.2em}
  \item Make a \emph{relative} comparison: even if both are weak, pick the better one unless they are truly indistinguishable.
  \item Use \textbf{Tie} when (a) differences are negligible, or (b) each is clearly better on different aspects of the \emph{same axis} and you cannot justify a preference.
  \item You may replay each audio response; judge based on the responses as presented (do not infer unstated content).
\end{itemize}

\subsection{Aggregation}
For each item and axis, we aggregate annotator choices by \textbf{majority vote}.
If no strict majority exists (e.g., near-even split across A/B/Tie), we assign the item outcome as \textbf{Tie} for that axis.
We then compute Win/Tie/Loss rates over items (Win: our model preferred; Loss: baseline preferred; Tie: no preference)~\cite{EN20120117X}.

\subsection{Statistical testing}
To test whether our model is preferred more often than chance, we perform a two-sided paired preference sign test for each axis.
We exclude Ties and let $W$ and $L$ denote the number of items where our model wins or loses, respectively ($N=W+L$).
Under the null hypothesis of no preference, wins follow a Binomial$(N, 0.5)$ distribution.
We report a two-sided $p$-value:

\begin{equation}
\resizebox{0.95\linewidth}{!}{$
p = 2 \cdot \min\Big( \Pr[\mathrm{Bin}(N,0.5) \ge W], \; \Pr[\mathrm{Bin}(N,0.5) \ge L] \Big).
$}
\end{equation}
We report the resulting Win/Tie/Loss percentages and $p$-values in Main Text.

% ---------------------------------
\section{EMA Sensitivity Analysis and Dynamic Weight Trajectory}
\label{sec:appendix_ema_sensitivity}

\subsection{Sensitivity Analysis of EMA Coefficient $\alpha$}

Table~\ref{tab:ema_sensitivity} reports IQ and EQ scores under different EMA coefficients $\alpha$, with group size $G=4$ on VITA-Audio.

\begin{table}[h]
\centering
\small
\begin{tabular}{lcc}
\toprule
\textbf{EMA Coefficient ($\alpha$)} & \textbf{IQ} & \textbf{EQ} \\
\midrule
$\alpha=0$ (no EMA) & 53.15 & 2.53 \\
$\alpha=0.5$ (low smoothing) & 54.80 & 2.85 \\
$\alpha=0.9$ (ours, default) & \textbf{55.24} & \textbf{2.92} \\
$\alpha=0.99$ (high smoothing) & 50.95 & 2.88 \\
\bottomrule
\end{tabular}
\caption{Sensitivity of IQ and EQ to the EMA coefficient $\alpha$.}
\label{tab:ema_sensitivity}
\end{table}

Under-smoothing ($\alpha=0.5$) leads to high variance in $\lambda_t$, destabilizing training updates. Over-smoothing ($\alpha=0.99$) introduces excessive lag: $\lambda_t$ fails to rise quickly enough even when rollout quality improves, causing the model to remain dominated by the SFT objective and missing opportunities for semantic refinement via preference optimization. Our default $\alpha=0.9$ provides the best balance.

We also report results with larger group size $G=8$:

\begin{table}[h]
\centering
\small
\begin{tabular}{lccc}
\toprule
\textbf{Group size ($G$)} & \textbf{EMA} & \textbf{IQ} & \textbf{EQ} \\
\midrule
4 & \ding{55} & 53.15 & 2.53 \\
4 & \ding{51} & 55.24 & 2.92 \\
8 & \ding{55} & 54.36 & 2.66 \\
8 & \ding{51} & \textbf{57.19} & 2.90 \\
\bottomrule
\end{tabular}
\caption{Effect of group size $G$ and EMA on IQ and EQ. IQ is mean over VoiceBench reasoning subsets; EQ is average VStyle score.}
\label{tab:group_size}
\end{table}

Increasing $G$ yields further IQ gains but brings little improvement in EQ. Removing EMA causes a substantial performance drop regardless of $G$, confirming that EMA stabilizes the effective mixing coefficient across steps rather than merely compensating for small-group noise. We set $G=4$ as it provides a strong trade-off between computational cost and overall performance.

\subsection{Dynamic Weight $\lambda_t$ Trajectory During Training}

We traced $\lambda_t$ throughout training and observed the following trend:

\textbf{Early training}: $\lambda_t$ starts low (typically $\lambda_t \approx 0.1$--$0.2$). The model's rollouts have high variance and lower reward reliability; the gating mechanism correctly suppresses the preference loss and relies more on SFT for acoustic anchoring.

\textbf{Mid-to-late training}: As the policy improves, rollout quality and discriminability increase. $\lambda_t$ gradually rises, allowing preference optimization to take a larger role in refining semantic intelligence.

\textbf{Convergence}: $\lambda_t$ stabilizes in the range of $[0.35, 0.55]$, confirming that the dynamic weight converges to a balanced regime where both SFT and preference optimization contribute, rather than collapsing to a single objective.

\end{document}